\documentclass[runningheads]{llncs}
\makeatletter
\newcommand{\@chapapp}{\relax}%
\makeatother
\usepackage{graphicx}

\usepackage{tikz}
\usepackage{comment}
\usepackage{amsmath,amssymb} 
\usepackage{color}

\usepackage[accsupp]{axessibility}  
\usepackage[backref]{hyperref}

\usepackage{enumitem}
\usepackage{booktabs}
\usepackage{makecell}
\usepackage{multirow}
\usepackage{colortbl}
\usepackage{bbding}
\usepackage{pifont}
\newcommand{\smark}{\textcolor{black}{\ding{51}}{\textcolor{black}{\kern-0.57em\ding{55}}}}
\newcommand{\cmark}{\ding{52}}
\newcommand{\xmark}{\ding{55}}
\usepackage{subfigure}
\usepackage{caption}
\usepackage{color}
\usepackage[marginal]{footmisc}

\usepackage{url}
\usepackage{capt-of}
\usepackage{appendix}

\begin{document}

\pagestyle{headings}
\mainmatter
\def\ECCVSubNumber{1792}  

\title{TO-Scene: A Large-scale Dataset for Understanding 3D Tabletop Scenes} 

\titlerunning{TO-Scene}
%
\footnotetext[1]{\label{footnote:equal}M. Xu and P. Chen contribute equally.}
\footnotetext[2]{\label{footnote:corr}Corresponding author.}
\author{Mutian Xu\inst{1}$^{\ref{footnote:equal}}$ \and
Pei Chen\inst{1}$^{\ref{footnote:equal}}$ \and
Haolin Liu\inst{1,2} \and
Xiaoguang Han\inst{1,2}$^{\ref{footnote:corr}}$}

\authorrunning{Xu et al.}
%
\institute{School of Science and Engineering, The Chinese University of Hong Kong, Shenzhen \and
The Future Network of Intelligence Institute, CUHK-Shenzhen\\
\email{\{mutianxu, peichen, haolinliu\}@link.cuhk.edu.cn, \{hanxiaoguang\}@cuhk.edu.cn}}

\maketitle

\begin{figure}
\centering
\includegraphics[width=\textwidth]{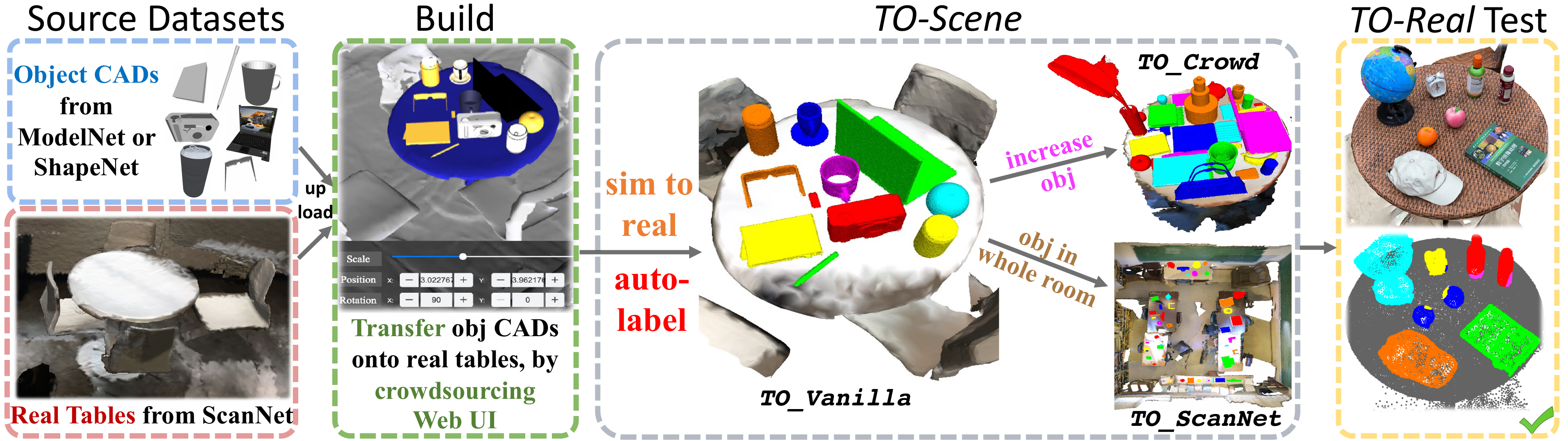}
\caption{Overview of data acquisition framework and TO-Scene dataset. We firstly transfer object CADs from ModelNet \cite{modelnet} and ShapeNet \cite{shapenet} onto real tables from ScanNet \cite{scannet} via crowdsourcing Web UI. Then  the tabletop scenes are simluated into real scans and annotated automatically. Three variants (TO\_Vanilla, TO\_Crowd and TO\_ScanNet) are presented for various scenarios. A real-scanned test data TO-Real is provided to verify the practical value of TO-Scene.}
\label{fig:intro}
\end{figure}

\begin{abstract}
Many basic indoor activities such as eating or writing are always conducted upon different \textit{tabletops} (e.g., coffee tables, writing desks). It is indispensable to understanding tabletop scenes in 3D indoor scene parsing applications. Unfortunately, it is hard to meet this demand by directly deploying data-driven algorithms, since 3D tabletop scenes are rarely available in current datasets. To remedy this defect, we introduce TO-Scene, a large-scale dataset focusing on \textbf{t}ablet\textbf{o}p scenes, which contains 20,740 scenes with three variants. To acquire the data, we design an efficient and scalable framework, where a crowdsourcing UI is developed to transfer CAD objects from ModelNet \cite{modelnet} and ShapeNet \cite{shapenet} onto tables from ScanNet \cite{scannet}, then the output tabletop scenes are simulated into real scans and annotated automatically.

Further, a tabletop-aware learning strategy is proposed for better perceiving the small-sized tabletop instances. Notably, we also provide a \textit{real} scanned test set TO-Real to verify the practical value of TO-Scene. Experiments show that the algorithms trained on TO-Scene indeed work on the realistic test data, and our proposed tabletop-aware learning strategy greatly improves the state-of-the-art results on both 3D semantic segmentation and object detection tasks. Dataset and code are available at \href{https://github.com/GAP-LAB-CUHK-SZ/TO-Scene}{https://github.com/GAP-LAB-CUHK-SZ/TO-Scene}.

\end{abstract}

\section{Introduction}
Understanding indoor scenes is a fundamental problem in many industrial applications, such as home automation, scene modeling, virtual reality, and perception assistance. While this topic spawns the recent development of 3D supervised deep learning methods \cite{paconv,pointtrans,groupfree}, their performance directly depends on the availability of large labeled training datasets. Thanks to the progress of 3D scanning technologies and depth sensors (e.g., Kinect \cite{kinect}), various 3D indoor scene datasets arised \cite{nyuv2,sls,sun3d,sunrgbd,scenenn,pigraphs,s3dis,scannet,Matterport3D}. 
The most popular indoor scene benchmark among them, ScanNet \cite{scannet}, is consisted of richly annotated RGB-D scans in real world, which is produced by a scalable data acquisition framework. 

Albeit the great advance on 3D indoor datasets allows us to train data-hungry algorithms for scene understanding tasks, one of the most frequent and widely-used setups under indoor environments is poorly investigated -- the scene focusing on \textit{tabletops}.



In indoor rooms, for satisfying the basic daily requirements (such as eating, writing, working), humans (or robots) need to frequently interact with or face to different tabletops (e.g., dining tables, coffee tables, kitchen cabinets, writing desks), and place, catch or use various tabletop objects (e.g., pencils, keyboards, mugs, phones, bowls). Thus, perceiving and understanding tabletop scenes is \textit{indispensable} in indoor scene parsing applications. Unfortunately, it is hard to meet this demand by directly deploying the 3D networks, since existing indoor scene datasets lack either adequate samples, categories or annotations of tabletop objects (illustrated in Table \ref{table:comparison}), from where the models is not able to learn the corresponding representations. Therefore, it is substantially meaningful to build a dataset focusing on the tabletop objects with sufficient quantities and classes.

We attempt to construct such a dataset and present our work by answering the below questions:

(\textbf{1}) \textit{Assumption -- What is the sample pattern supposed to be?} Starting from our previous motivation, the dataset is expected to meet the practical demand in real applications. Thus, the sample is assumed to look similar with the real scanned tabletop scene, which is a bounded space filled by a table with multiple objects above it, and surrounded by some background furniture.
This requires the model to perceive both the individual objects and their inter-relationships, while assisted by the context information learned from finite indoor surroundings.

(\textbf{2}) \textit{Acquisition -- How to build such a dataset with decent richness, diversity, and scalability, at a low cost?} Building large-scale datasets is always challenging, not only because of the laborious collection of large amounts of 3D data, but also the non-trivial annotation.
As for our work, it is undoubtedly burdensome to manually place objects above real tables, then scan and label 3D data. Instead, we design an efficient and scalable framework (Fig. \ref{fig:intro}) to overcome this difficulty.
We firstly develop a Web UI (Fig. \ref{fig:UI}) to help novices transfer CAD objects of ModelNet \cite{modelnet} and ShapeNet \cite{shapenet} to suitable tables extracted from ScanNet \cite{scannet} scenes. The UI makes it possible to enlarge our dataset in the future.
After that, we simulate the synthetic objects into real-world data, expecting that the model trained on our dataset can work on real scanned data. 
Last, an automatic annotation strategy is adopted to produce point-wise semantic labels on each reconstructed object's meshes, based on its bounding box that directly gained from the CAD model. So far, the complete acquisition pipeline enables us to construct a vanilla dataset as mentioned in (\textbf{1}), which contains 12,078 tabletop scenes with 60,174 tabletop instances belonging to 52 classes, called \textbf{TO\_Vanilla}.

(\textbf{3}) \textit{Enrichment -- Can we create more variants of the data to bring new challenges into the indoor scene parsing task?}
i) In our daily life, the tabletops are sometimes full of crowded objects. 
To simulate this situation, we increase the instances above each table, 
producing a more challenging setup called \textbf{TO\_Crowd}, which provides 3,999 tabletop scenes and 52,055 instances that are distinguished with TO\_Vanilla.
ii) Both TO\_Vanilla and TO\_Crowd assume to parse the tabletop scenes. Nevertheless, some real applications require to parse the \textit{whole} room with all furniture including tabletop objects in one stage. To remedy this, another variants \textbf{TO\_ScanNet} comes by directly using the tables in TO-Vanilla, but the complete scans of rooms that accommodate the corresponding tables are still kept. It covers 4663 scans holding around 137k tabletop instances, which can be treated as an \textit{augmented ScanNet} \cite{scannet}. Combining three variants, we introduce \textbf{TO-Scene}.

(\textbf{4}) \textit{Strategies -- How to handle the open challenges in our TO-Scene?}
The tabletop objects in TO-Scene are mostly in smaller-size compared with other large-size background furniture, causing challenges to discriminate them. 
To better perceive the presence of tabletop instances, we propose a tabletop-aware learning strategy that can significantly improve upon the state-of-the-art results on our dataset, by jointly optimizing a tabletop-object discriminator and the main segmentation or detection targets in a single neural network.

(\textbf{5}) \textit{Practicality -- Can TO-Scene indeed serve for real applications?} To investigate this, we manually scan and annotate three sets of data, that corresponds to the three variants of TO-Scene. We denote the whole test data as \textbf{TO-Real}, which provides 197 \textit{real} tabletop scans with 2373 objects and 22 indoor room scans holding 824 instances.
Consequently, the models trained on TO-Scene get promising results on our realistic test data, which suggests the practical value of TO-Scene.

Here, the contributions of this paper are summarized as:
\begin{itemize}[noitemsep,topsep=0pt]
	\item[$\bullet$] TO-Scene -- To the best of our knowledge, the first large-scale dataset primarily for understanding tabletop scenes, with three different variants.
	\item[$\bullet$] An efficient and scalable data acquisition framework with an easy-to-use Web UI for untrained users.
	\item[$\bullet$] A tabletop-aware learning strategy, for better discriminating the small-sized tabletop instances in indoor rooms. 
	\item[$\bullet$] A real scanned test set -- TO-Real, with the same three variants as TO-Scene, for verifying the practical value of TO-Scene.
	\item[$\bullet$]Experiments demonstrate that the networks running on TO-Scene work well on TO-Real, and our proposed tabletop-aware learning strategy greatly improves the state-of-the-arts.
	\item[$\bullet$] TO-Scene and TO-Real, plus Web UI are all open source.
\end{itemize}


\setlength{\tabcolsep}{2pt}
\begin{table}[t]
\begin{center}
\caption{Overview of 3D indoor scene datasets. ``kit'' indicates kitchen, ``obj'' denotes object, ``bbox'' means bounding boxes, ``point segs'' is point-wise segmentation. Our large-scale TO-Scene with three variants focuses on tabletop scenes, and is efficiently built by crowdsourcing UI and automatic annotation.}
\label{table:comparison}
\resizebox{1.0\textwidth}{!}{
\begin{tabular}{l|cccc}
\Xhline{1.0pt}
\noalign{\smallskip}
Dataset & Tabletop & $\sharp$Scenes & Collection & Annotation\\
\noalign{\smallskip}
\Xhline{0.6pt}
\noalign{\smallskip}
SUNCG \cite{suncg} & \xmark & 45k & synthetic, by designers &  dense 3D\\
\noalign{\smallskip}
\hline
\noalign{\smallskip}
SceneNet \cite{scenenet} & \xmark & 57 & synthetic, by designers & dense 3D\\
\noalign{\smallskip}
\hline
\noalign{\smallskip}
OpenRooms \cite{openroom} & \xmark & 1068 & synthetic, by designers & dense 3D\\
\noalign{\smallskip}
\hline
\noalign{\smallskip}
3D-FRONT \cite{front} & \xmark & 19k & synthetic, by designers & dense 3D\\
\noalign{\smallskip}
\Xhline{0.6pt}
\noalign{\smallskip}
NYU v2 \cite{nyuv2} & \xmark & 464 & scan, by experts & raw RGB-D \cite{labelme}\\
\noalign{\smallskip}
\hline
\noalign{\smallskip}
SUN 3D \cite{sun3d} & \xmark & 415 & scan, by experts & 2D polygons\\
\noalign{\smallskip}
\hline
\noalign{\smallskip}
S3DIS \cite{s3dis} & \xmark & 265 & scan, by experts & dense 3D \cite{cloudcompare}\\
\noalign{\smallskip}
\hline
\noalign{\smallskip}
ScanNet \cite{scannet} & \xmark & 1513 & scan, by crowdsourcing UI \cite{scannet} & dense 3D\\
\noalign{\smallskip}
\Xhline{0.6pt}
\noalign{\smallskip}
WRGB-D \cite{close} & \smark, 5 sorts small obj & 22 & scan, by experts & point segs, 2D polygons\\
\noalign{\smallskip}
\hline
\noalign{\smallskip}
GMU Kit \cite{kitchen} & \smark, 23 sorts kit obj & 9 & scan, by experts & dense 3D\\
\noalign{\smallskip}
\hline
\noalign{\smallskip}
\rowcolor{gray!20}
\textbf{TO-Scene} & \textbf{\cmark,} & \textbf{21k,} & \textbf{effortless transfer} & \textbf{dense 3D:}\\
\rowcolor{gray!20}
\textbf{(ours)} & \textbf{various tables,} & \textbf{3 variants with} & 
by our-own & \textbf{bboxes of \cite{shapenet,modelnet}} \\
\rowcolor{gray!20}
 & \textbf{52 sorts obj} & \textbf{augmented \cite{scannet}} &  \textbf{crowdsourcing UI} & \textbf{ + auto point segs}\\
\noalign{\smallskip}
\Xhline{1.0pt}
\end{tabular}
}
\end{center}
\end{table}
\setlength{\tabcolsep}{2pt}

\section{Related Work}
\noindent{\textbf{3D indoor scene datasets.}}
3D indoor scene datasets have been actively made over the past few years. NYU v2 \cite{nyuv2} is an early real dataset for RGBD scene understanding, which contains 464 short RGB-D videos captured from 1449 frames, with 2D polygons annotations as LabelMe \cite{labelme} system.  
SUN3D \cite{sun3d} captures a set of 415 sequences of 254 spaces, with 2D polygons annotation on key frames. The following SUN RGB-D \cite{sunrgbd} collects 10,335 RGB-D frames with diverse scenes. Yet it does not provide complete 3D surface reconstructions or dense 3D semantic segmentations. To remedy this, Hua et al. \cite{scenenn} introduce SceneNN, a larger scale RGB-D dataset consisting of 100 scenes. 
Another popular dataset S3DIS \cite{s3dis} includes manually labeled 3D meshes for 265 rooms captured with a Matterport camera. Later, Dai et al. \cite{scannet} design an easy-to-use and scalable RGB-D capture system to produce ScanNet, the most widely-used and richly annotated indoor scene dataset, which contains 1513 RGB-D scans of over 707 indoor rooms with estimated camera parameters, surface reconstructions, textured meshes, semantic segmentations. Further, Matterport3D \cite{Matterport3D} provides 10,800 panoramic views from 194,400 RGB-D images of 90 building scenes. In \cite{suncg,scenenet,openroom,front}, the synthetic 3D indoor scene data are generated. The recent 3D-FRONT \cite{front} contains 18,797 rooms diversely furnished by 3D objects, surpassing all public scene datasets.

The aforementioned datasets ignore an important data form in indoor scene parsing applications -- \textit{tabletop} scenes, which is the basis of our dataset. There are two existing datasets emphasizing small objects that may appear on tables. WRGB-D Scenes \cite{close} includes 22 annotated scene video sequences containing 5 sorts of small objects (subset of WRGB-D Object \cite{close}). GMU kitchen \cite{kitchen} is comprised of multi-view kitchen counter-top scenes, each containing 10-15 hand-hold instances annotated with bounding boxes and in the 3D point cloud. However, their complex data collections and annotations cause the severe limitation on the data richness, diversity and scalability (see Table \ref{table:comparison}). 

\noindent{\textbf{3D shape and object datasets.}}
The tabletop objects in TO-Scene are originated from ModelNet \cite{modelnet} containing 151,128 3D CAD models of 660 categories, and ShapeNetCore \cite{shapenet} covering 55 object classes with 51,300 3D models.
As for 3D shape datasets, \cite{shapenet,modelnet,abc,scan2cad,objnet,pascal3d} provide CAD models, while \cite{redwood,bigbird,photoshape,scanobjectnn,abo} advocate the realistic data. \cite{close,co3d} contain multi-view images of 3D objects, and the recent Objectron \cite{objectron} is a collection of short object-centric videos.
The aforesaid datasets deal with single 3D objects. In contrast, our TO-Scene highlights the holistic understanding and relationship exploration on various tabletop objects under indoor room environments.

\noindent{\textbf{Robot grasping and interacting datasets.}}
\cite{cornellgrasp,jacquard,graspnet} contribute to the grasping of tabletop objects. Their annotations of object 6D poses and grasp poses exclude object categories, which are customized for the robot grasping, instead of understanding tabletop scenes. Garcia et al. \cite{robotrix} introduce a hyperrealistic indoor scene dataset, which is explored by robots interacting with objects in a simulated world, but still specifically for robotic vision tasks.

\noindent{\textbf{3D indoor scene parsing algorithms.}}
The bloom of the indoor scene datasets opens the chances for training and benchmarking deep neural models in different 3D scene understanding tasks, such as 3D semantic segmentation \cite{pointnet2,gdanet,paconv,pointtrans} and 3D object detection \cite{votenet,h3dnet,groupfree}. 
Our TO-Scene raises a new challenge of discriminating the small-sized tabletop objects in indoor rooms. To tackle this, we propose a tabletop-aware learning strategy with a jointly optimized tabletop-object discriminator, for better perceiving tabletop instances.

\section{Data Acquisition}
Building large-scale 3D datasets is always challenging, not only because of the laborious collection of large amounts of 3D data, but also the non-trivial annotation. This problem is especially severe even for expert users when collecting cluttered tabletop scenes under indoor environments, as in \cite{kitchen,close}, which makes it non-trivial to replenish the dataset (i.e., inferior scalablity), and limits the richness and diversity of the dataset.

To solve this issue, we present an efficient and scalable framework for creating TO-Scene, with the ultimate goal of driving the deep networks to have a significant effect on the real world 3D data.

\begin{figure}[t]
\centering
\includegraphics[width=0.6\textwidth]{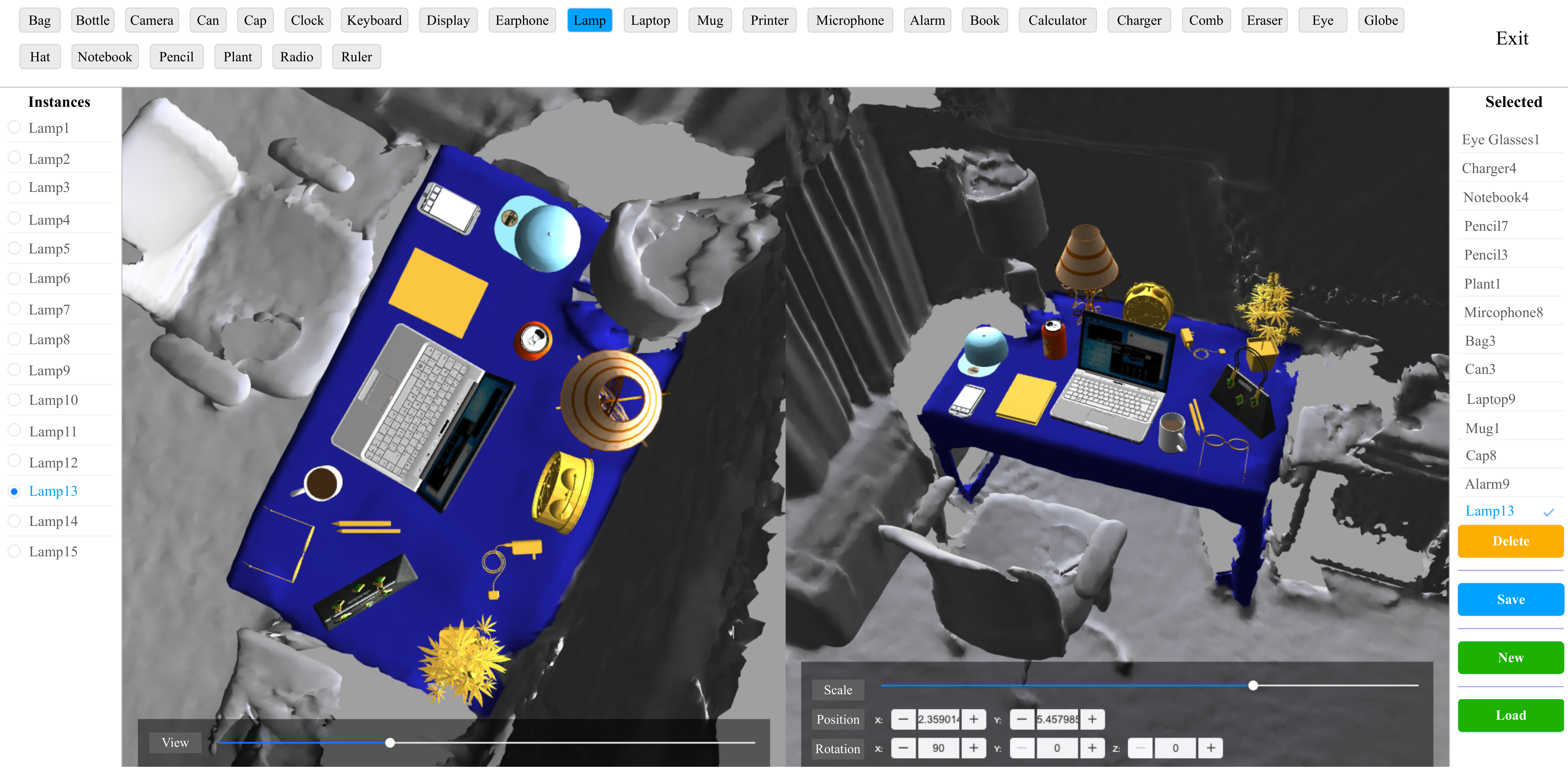}
\caption{Our web-based crowdsourcing interface for transferring CADs onto real tabletops. The user chooses an suitable object and click somewhere on Bird's-Eye-View (BEV) (left) for placing it above 3D tables (right).}
\label{fig:UI}
\end{figure}

\subsection{Transfer Object CADs into Indoor Scenes}
In this work, rather than manually placing and scanning the tabletop objects in the real world, we propose to transfer the object CADs from ModelNet \cite{modelnet} and ShapeNet \cite{shapenet} into the tables located in ScanNet \cite{scannet} rooms.~\\

\noindent{\textbf{Source datasets.}}
ShapeNetCore \cite{shapenet} contains 55 object classes with 51,300 3D models and organizes them under the WordNet \cite{wordnet} taxonomy. Despite the tremendous size of ShapeNetCore, its quantity of tabletop objects are not able to meet our requirement. Thus, we also employ ModelNet \cite{modelnet} covering 151,128 3D CAD models of 660 classes. Both of them have the annotations of consistent rigid alignments and physical sizes. Leveraging their richness, we are able to agreeably borrow a large amount of CAD instances, belonging to 52 classes (detailed in Fig. \ref{fig:to1_stat}) that are commonly seen on different tables in our daily life, as the tabletop objects of TO-Scene.

Another important source is ScanNet \cite{scannet}, which is a richly annotated real-world indoor scene dataset including 2.5M RGB-D images in 1513 scans acquired in 707 distinct spaces, especially covering diverse table instances. We extract the tables from ScanNet to place tabletop objects. Additionally, since the scanning of a tabletop scenes in real applications will also cover some background furniture, the ScanNet indoor environments around the tables also act as the context in our tabletop scenes.~\\ 

\noindent{\textbf{User interface.}}
For allowing untrained users to create large-scale datasets, we develop a trivial-to-use Web UI to ease the object transfer process. As shown in Fig. \ref{fig:UI}, the Web UI shows up with a BEV (Bird's-Eye-View) of a table on the left and a 3D view of the surroundings on the right. The table are randomly selected from ScanNet with CADs picked from ModelNet and ShapeNet each time. The operation is friendly for untrained users. Specifically, when placing an object, the user does \textit{not} need to perceive the \textit{height} for placing it in 3D. The only action is just clicking at somewhere on the table in BEV, then the selected object will be automatically placed on the position as the user suggests, based on an encapsulated precise calibration algorithm. The ﬂow of data through the system is handled by a tracking server. More details of UI can be found in the supplementary material.

The UI makes it possible to enlarge our dataset, endowing TO-Scene with decent scalability. Besides, similar transfer between different types of datasets can be achieved by revising the UI for creating new datasets or various purposes.~\\

\noindent{\textbf{Transfer rules.}}
Notably, two implicit rules during transfer are achieved:

(1) \textit{Rationality}. The function of a table is supposed to \textit{match} with the objects above it. For instance, a mug will more likely to be appeared on coffee tables, while a pencil is possibly placed on writing desks. To achieve this, we set the UI to present objects only from the categories that fit the selected table, and the table label as well as its surroundings are shown to help users figure out the table function. Besides, users are guided to place objects according to their commonsense knowledge instead of reckless operations.

(2) \textit{Diversity}. As mentioned in (1), the tables are randomly picked by UI with various instances shown to fit the table functions. Additionally, we employ around 500 users who may face with diverse tabletop scenes in their daily life, from different professions (e.g., teacher, doctor, student, cook, babysitter) and ages (20$\sim$50). After they finish, about 200 new users will double-check the samples, when they can rearrange, add or delete the objects.~\\

\noindent{\textbf{Storage.}}
Each time the user finishes a transfer, the UI will generate a file recording the file names of CADs and rooms, table ID, as well as the calibration parameters. This makes our dataset parameterized and editable, promoting us to store the results in a memory-saving way.

\subsection{Simulate Tabletop Scenes into Real Scans}
We realize a possible domain mismatch between the original synthetic tabletop objects and the real scanned data, which may result that the deep algorithms trained on TO-Scene are not able to directly work on real-world data. To avoid this, it is necessary to simulate the CAD objects into realistic data.

According to the real-world data collection procedures, we firstly utilize Blender \cite{blender,blenderproc} to render CAD objects into several RGB-D sequences, via emulating the real depth camera. For instance, when a table is against walls, our simulated camera poses will only cover the views in front of tables.
Generally, the objects are visible from different viewpoints and distances and may be partially or completely occluded in some frames. Then, the rendered RGB-D frames are sent to TSDF \cite{tsdf} reconstruction algorithms for generating the realistic 3D data.

So far, the whole process brings the occlusions and reconstruction errors, simulating the noise characteristics in real data scanned by depth sensors. As a matter of course, the data domain of tabletop objects are \textit{unified} with the tables and background furniture that are extracted from real scanned ScanNet \cite{scannet}. Table \ref{table:real_test} demonstrates that the model trained on our dataset can be generalized to realistic data for practical applications.


\subsection{Annotate Tabletop Instances}
Agreeably, the bounding box annotations (i.e., the center coordinates of boxes, length, width, height, and semantic label) of tabletop objects are naturally gained from their CAD counterparts. Next, since the bounding box of an instance delineates an area covering its owning points, this promotes us to get the point-wise annotations straightforwardly and automatically.

\begin{figure}[t]
\centering
\includegraphics[width=\textwidth]{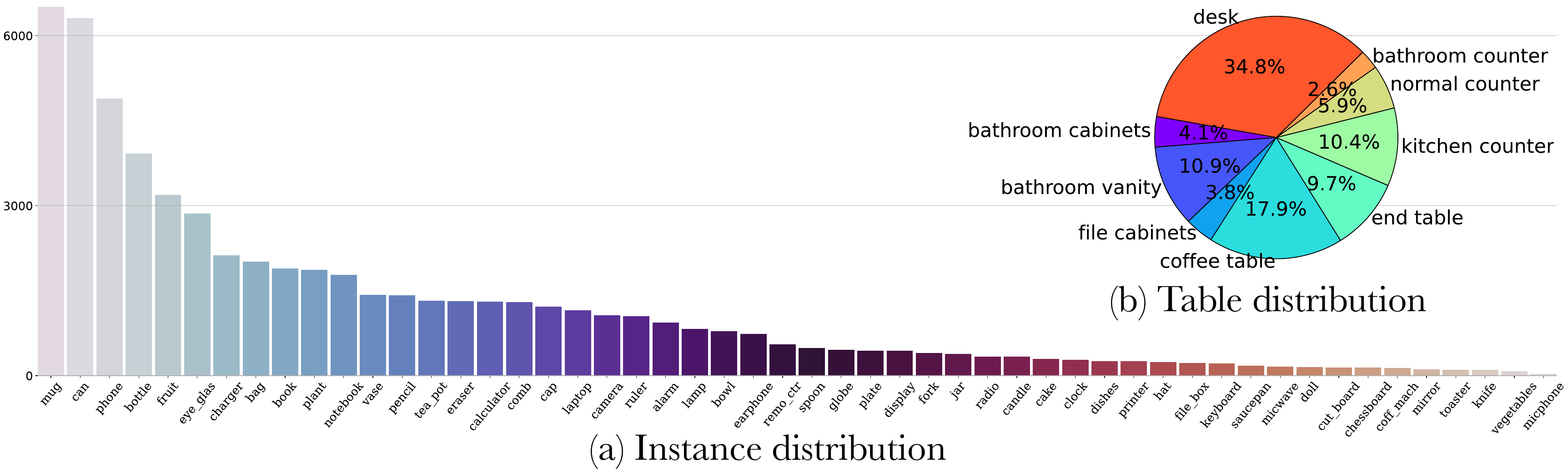}
\caption{Semantic annotation statistics of TO\_Vanilla.}
\label{fig:to1_stat}
\end{figure}

Following above procedures, we build a dataset consisting of 12,078 tabletop scenes, with 60,174 tabletop object instances from 52 common classes.
Fig. \ref{fig:sample} (a)) shows a sample of this vanilla dataset, denoted as \textbf{TO\_Vanilla}, which is also the foundation of our dataset. Fig. \ref{fig:to1_stat} shows the statistics for the semantic annotation of the major tabletop objects and the used table categories in TO\_Vanilla.

\subsection{Data Enrichment}
We construct another two variants upon TO\_Vanilla, for benchmarking existing algorithms under more real scenarios with new challenges.~\\

\noindent{\textbf{Crowded objects.}} 
The tabletops in our daily life are often full of crowded objects with more inter-occlusions, 
To simulate this challenge, we reload some object-table sudo-mappings outputs when building TO\_Vanilla (i.e., the tabletops after object transfers, yet not rendered or reconstructed), and employ novices to place much more tabletop CADs above each table. Then we render and reconstruct the new crowded set via the same way as before.
Consequently, more occlusions with reconstruction flaws are introduced (see Fig. \ref{fig:sample} (b)), yielding a more challenging setup of tabletop scenes, indicated by \textbf{TO\_Crowd}. It covers 3,999 tabletop scenes and 52,055 instances.~\\

\noindent{\textbf{Parse whole room in one stage.}} 
Previous TO\_Vanilla and TO\_Crowd assume to only parse tabletop scenes, but many real applications require to process the \textit{whole} room with all furniture including tabletop objects in one stage.
To make up this situation, we keep the complete scans of ScanNet~\cite{scannet} rooms in each data sample, from which the tables of TO\_Vanilla are extracted. We maintain the semantic label on original room furniture from ScanNet. As a result, another variant \textbf{TO\_ScanNet} is presented (see Fig.~\ref{fig:sample} (c)), which requires algorithms to comprehensively understand both tabletop objects and room furniture. TO\_ScanNet can be regarded as an \textit{augmented ScanNet}.~\\

\subsection{\textbf{TO-Scene Dataset.}} 
Finally, \textbf{TO-Scene} is born by combining TO\_Vanilla, TO\_Crowd and TO\_ScanNet. The train/validation split statistics of three variants are summarized in Table \ref{table:split}, with the per-category details shown in the supplementary material. Our TO-Scene contains totally 16,077 tabletop scenes with 52 common classes of tabletop objects. The annotation includes vertex semantic labels, 3D bounding boxes, and camera poses during rendering, which opens new opportunities for exploring the methods in various real-world scenarios.

\begin{figure}[t]
\parbox[t]{0.49\linewidth}{\null
  \centering
  \includegraphics[height=2.5cm]{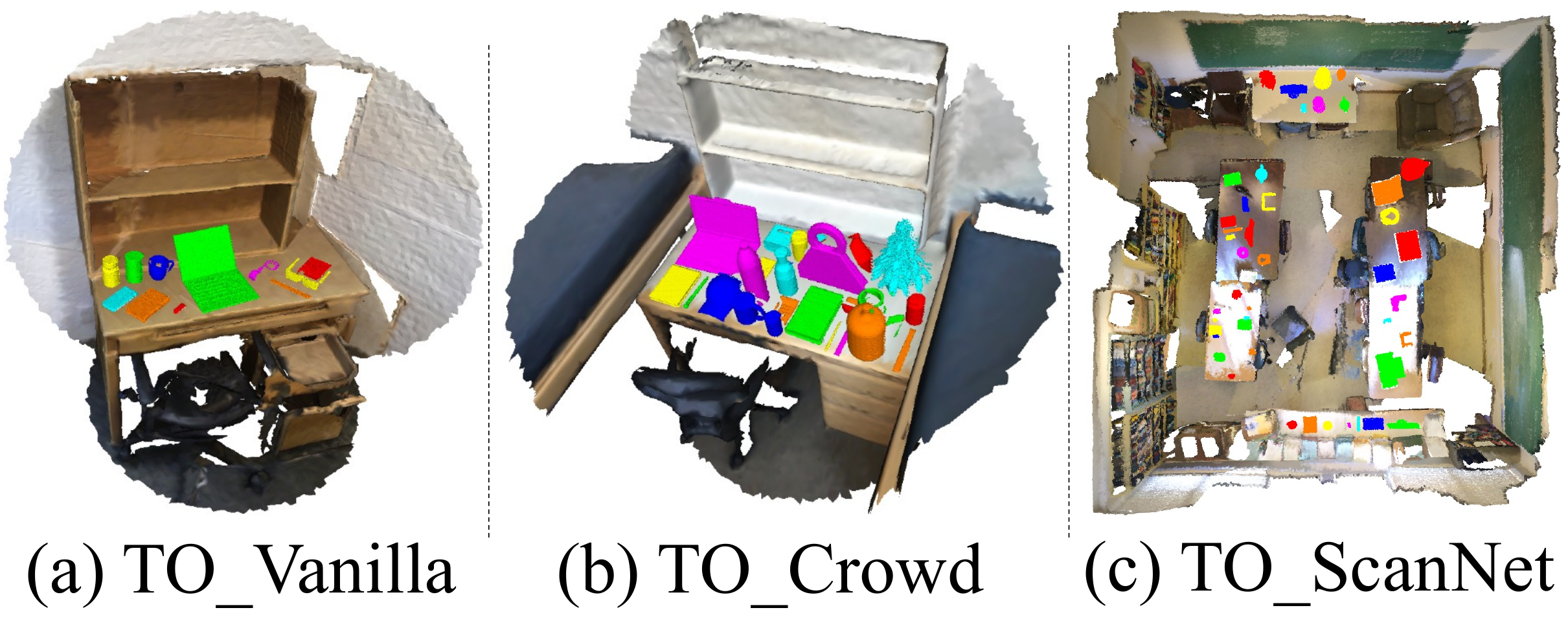}
  \captionof{figure}{Three variants in TO-Scene.}
  \label{fig:sample}
}
\parbox[t]{0.49\linewidth}{\null
\centering
  \vskip-\abovecaptionskip
  \captionof{table}[t]{Train/Test Split.}
  \label{table:split}
  \vskip\abovecaptionskip
    {\resizebox{0.4\textwidth}{!}{
\begin{tabular}{c|cc|cc}
\Xhline{1.0pt}
\noalign{\smallskip}
 & \multicolumn{2}{c|}{Scenes} & \multicolumn{2}{c}{Instances}\\
 & $\sharp$Train & $\sharp$Test & $\sharp$Train & $\sharp$Test \\
\noalign{\smallskip}
\Xhline{0.6pt}
\noalign{\smallskip}
TO\_Vanilla & 10003 & 2075 & 49982 & 10192\\
\noalign{\smallskip}
TO\_Crowd & 3350 & 649 & 43390 & 8665\\
\noalign{\smallskip}
TO\_ScanNet & 3852 & 811 & 114631 & 23032\\
\noalign{\smallskip}
\Xhline{1.0pt}
\end{tabular}
}}
}
\end{figure}

\textbf{Note:} The stated three variants are currently organized separately for different uses. One may combine either of them to explore more research prospects. Furthermore, this paper just presents the current TO-Scene snapshot, we will keep replenishing our dataset and provide extra \textit{private test set} for benchmarking in the future. More data samples can be found in the supplementary material.

\section{Tabletop-aware Learning Strategy}
For demonstrating the value of TO-Scene data, we focus on 3D semantic segmentation and 3D object detection tasks, towards understanding scenes from both point-wise and instance-level.

\begin{figure}[t]
\begin{minipage}[t]{0.4\linewidth}
\centering
\includegraphics[height=2.5cm]{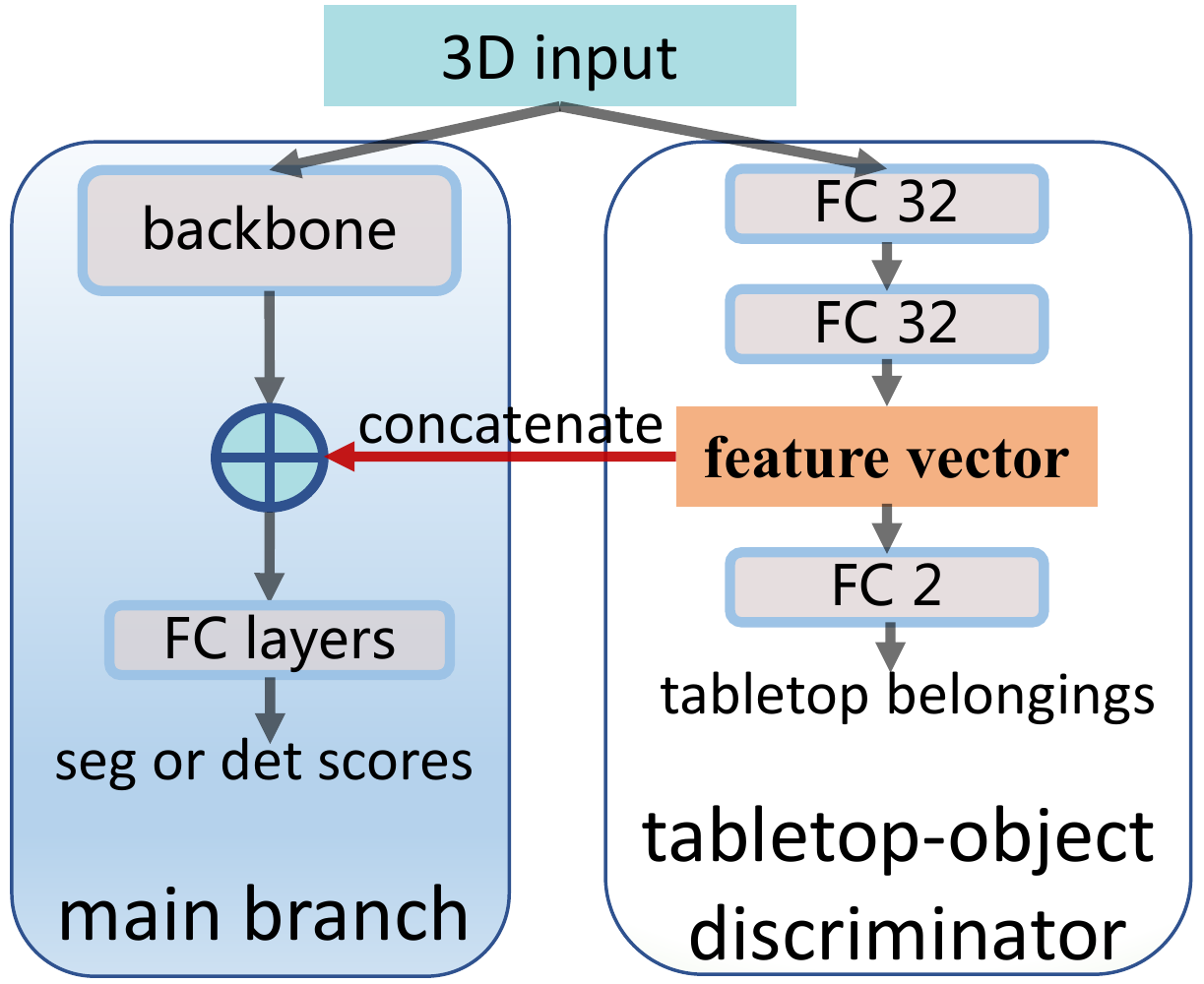}
\caption{Tabletop-aware learning.}
\label{fig:bin_feature}
\end{minipage}
\begin{minipage}[t]{0.55\linewidth}
\centering
\includegraphics[height=2.5cm]{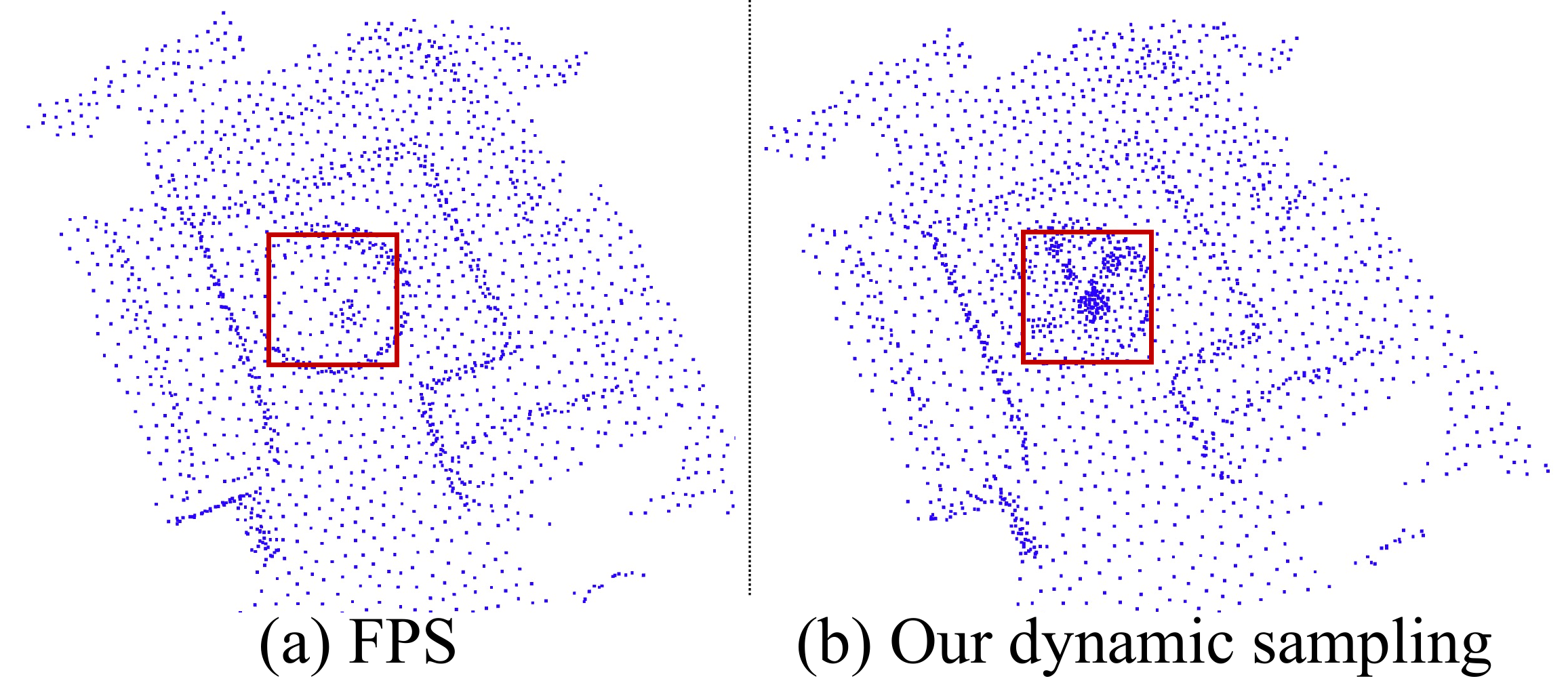}
\caption{FPS V.S. dynamic sampling.}
\label{fig:downsampling}
\end{minipage}
\end{figure}

In TO-Scene, since tabletop objects are mostly in smaller-size compared with other large-size background furniture, it is naturally difficult to discriminate tabletop instances, especially for TO\_ScanNet with lots of big furniture. Additionally, existing 3D networks mostly apply conventional downsampling such as farthest point sampling and grid sampling to enlarge the receptive field. Nevertheless, after being sampled by these schemes, the point densities of small tabletop objects are conspicuously sparser than big furniture (see Fig. \ref{fig:downsampling} (a)), which hurts the perceiving of tabletop objects. 

To handle these issues, our idea is to guide the network aware of the presence of tabletop objects, via adding a tabletop-object discriminator that is essentially a binary classifier. The loss is jointly optimized as the sum of the tabletop-object discriminator and segmentation (or detection) loss, which can be written as: $L_{total} = L_{seg\,or\,det} + \lambda L_{dis} $, where $\lambda$ is the weight. For segmentation, the $0-1$ ground truth (gt) for $L_{dis}$ can be directly gained from the point-wise semantic labels indicating if a point class belongs to tabletop objects, and $L_{dis}$ is a cross-entropy loss.
Yet the gt for detection only comes from bounding boxes. For fairly learning the discriminator, we employ a \textit{soft} point-wise gt label that is a normalized per-point distance between each point and the center of a gt tabletop object bounding box, and compute the mean squared error as $L_{dis}$.

In this work, two operations are derived by tabletop-object discriminator:

(1) As shown in Fig. \ref{fig:bin_feature}, the feature vector before the last fully connected layer of tabletop-object discriminator is concatenated with the features extracted by the main segmentation or detection branch, so that the predictions of the main branch are driven by the tabletop belongings information. 

(2) A dynamic sampling strategy is proposed, where the points with higher tabletop-object discriminator score (i.e., points of tabletop objects) are more likely to be sampled (see Fig. \ref{fig:downsampling} (b)). We replace the original sampling during the feature extraction in all backbone networks with our dynamic sampling.

In the practice, tabletop-object discriminator is implemented through a few fully-connected layers, assisted via max-pooling on K Nearest Neighbor (KNN) point features for fusing contextual information. Our joint learning concept can be promoted for tackling similar problems that requires to parse the objects with large size variance. The experimental results are listed in Sec.~\ref{sec:benchmark}. 


\section{Benchmark on TO-Scene}\label{sec:benchmark}
For making the conclusions solid, extensive experiments are conducted on 3D semantic segmentation and 3D object detection tasks.

\begin{table}[t]
\centering
\begin{minipage}[t]{0.5\linewidth}
\caption{Segmentation mIoU (\%).} 
\label{table:benchmark_seg}
\resizebox{1.0\textwidth}{!}{
\begin{tabular}{l|c|c|c}
\Xhline{1.0pt}
\noalign{\smallskip}
Method & TO\_Vanilla & TO\_Crowd & TO\_ScanNet\\
\noalign{\smallskip}
\Xhline{0.6pt}
\noalign{\smallskip}
PointNet \cite{pointnet} & 49.31 & 44.89 & 36.74\\
\noalign{\smallskip}
\hline
\noalign{\smallskip}
PointNet$^{++}$ \cite{pointnet2} & 65.57 & 61.09 & 53.97\\
PointNet$^{++}$ + FV & 68.74 & 64.95 & 57.23\\
PointNet$^{++}$ + DS & 67.52 & 63.28 & 56.97\\
PointNet$^{++}$ + FV + DS & 69.87 & 65.15 & 58.80\\
\noalign{\smallskip}
\hline
\noalign{\smallskip}
PAConv \cite{paconv} & 75.68 & 71.28 & 65.15\\
\noalign{\smallskip}
\hline
\noalign{\smallskip}
Point Trans \cite{pointtrans} & 77.08 & 72.95 & 67.17\\
Point Trans + FV & 79.01 & 75.06 & 69.09\\
Point Trans + DS & 77.84 & 73.81 & 68.34\\
Point Trans + FV + DS & \textbf{79.91} & \textbf{75.93} & \textbf{69.59}\\
\noalign{\smallskip}
\Xhline{1.0pt}
\end{tabular}}
\end{minipage}
\begin{minipage}[t]{0.49\linewidth}
\caption{Detection mAP@0.25 (\%).}
\label{table:benchmark_det}
\resizebox{1.0\textwidth}{!}{
\begin{tabular}{l|c|c|c}
\Xhline{1.0pt}
\noalign{\smallskip}
Method & TO\_Vanilla & TO\_Crowd & TO\_ScanNet\\
\noalign{\smallskip}
\Xhline{0.6pt}
\noalign{\smallskip}
VoteNet \cite{votenet} & 53.05 & 48.27 & 43.70\\
VoteNet + FV & 59.33 & 50.32 & 48.36\\
VoteNet + DS & 58.87 & 58.05 & 52.92\\
VoteNet + FV + DS & 60.06 & 58.87 & 56.93\\
\noalign{\smallskip}
\hline
\noalign{\smallskip}
H3DNet\cite{h3dnet} & 59.64 & 57.25 & 52.39\\
\noalign{\smallskip}
\hline
\noalign{\smallskip}
Group-Free 3D \cite{groupfree} & 61.75 & 59.61 & 49.04\\
Group-Free 3D + FV & 62.26 & 59.63 & 53.66\\
Group-Free 3D + DS & 62.19 & 59.69 & 55.71\\
Group-Free 3D + FV + DS & \textbf{62.41} & \textbf{59.76} & \textbf{57.57}\\
\noalign{\smallskip}
\Xhline{1.0pt}
\end{tabular}}
\end{minipage}
\end{table}

\subsection{Common Notes}
Below are common notes for both two tasks. The implementation details can be found in the supplementary material.


(1) We follow the original training strategies and data augmentations of all tested methods from their papers or open repositories. 

(2) In Table~\ref{table:benchmark_seg} and Table~\ref{table:benchmark_det}, ``FV'' means applying feature vector of tabletop-object discriminator, ``DS'' indicates our dynamic sampling strategy.

\subsection{3D Semantic Segmentation}

\noindent{\textbf{Pre-voxelization.}}
A common task on 3D data is semantic segmentation (i.e. labeling points with semantic classes). We advocate to pre-voxelize point clouds, which brings more regular structure and context information, as in \cite{sparseconv,mink,closerlook,kpconv,pointtrans}. 
We set the voxel size to $4mm^3$ for matching the small sizes of tabletop objects. After voxelization, every voxel stores a surface point with object class annotation. Then we randomly sample 80,000 points from all voxels in a scene for training, and all points are adopted for testing.~\\

\noindent{\textbf{Networks.}}
We benchmark PointNet \cite{pointnet}, PointNet++ \cite{pointnet2}, PAConv \cite{paconv} and Point Transformer \cite{pointtrans}.
PointNet++ and Point Transformer are chosen as the backbones to plug our tabletop-aware learning modules.~\\

\noindent{\textbf{Results and analysis.}}
Following the state-of-the-arts \cite{bpnet,paconv,pointtrans}, we use mean of classwise intersection over union (mIoU) as the evaluation metrics. As we can see from Table \ref{table:benchmark_seg}, the state-of-the-arts learned from the training data are able to perform well in the test set based on the geometric input. Moreover, our tabletop-aware learning modules stably improve the model performance, especially when they are applied together.

\subsection{3D Object Detection}
\noindent{\textbf{Data-preprocessing.}}
Understanding indoor scenes at the instance level is also important. 
We follow the original pre-processing schemes of the selected state-of-the-arts.~\\

\noindent{\textbf{Networks.}}
We run VoteNet \cite{votenet}, H3DNet \cite{h3dnet} and Group-Free 3D \cite{groupfree} on TO-Scene. VoteNet and Group-Free 3D are picked as the backbones to integrate our tabletop-aware learning strategies.~\\

\noindent{\textbf{Results and analysis.}}
For the evaluation metrics, we use mAP@0.25, mean of average precision across all semantic classes with 3D IoU threshold 0.25, following the state-of-the-arts.
As shown in Table \ref{table:benchmark_det}, the deep networks achieve good results based on the geometric input. Our tabletop-aware learning methods again significantly improve the model performance.


\begin{figure}[t]
\centering
\begin{minipage}[t]{0.62\linewidth}
\centering
\includegraphics[height=2.9cm]{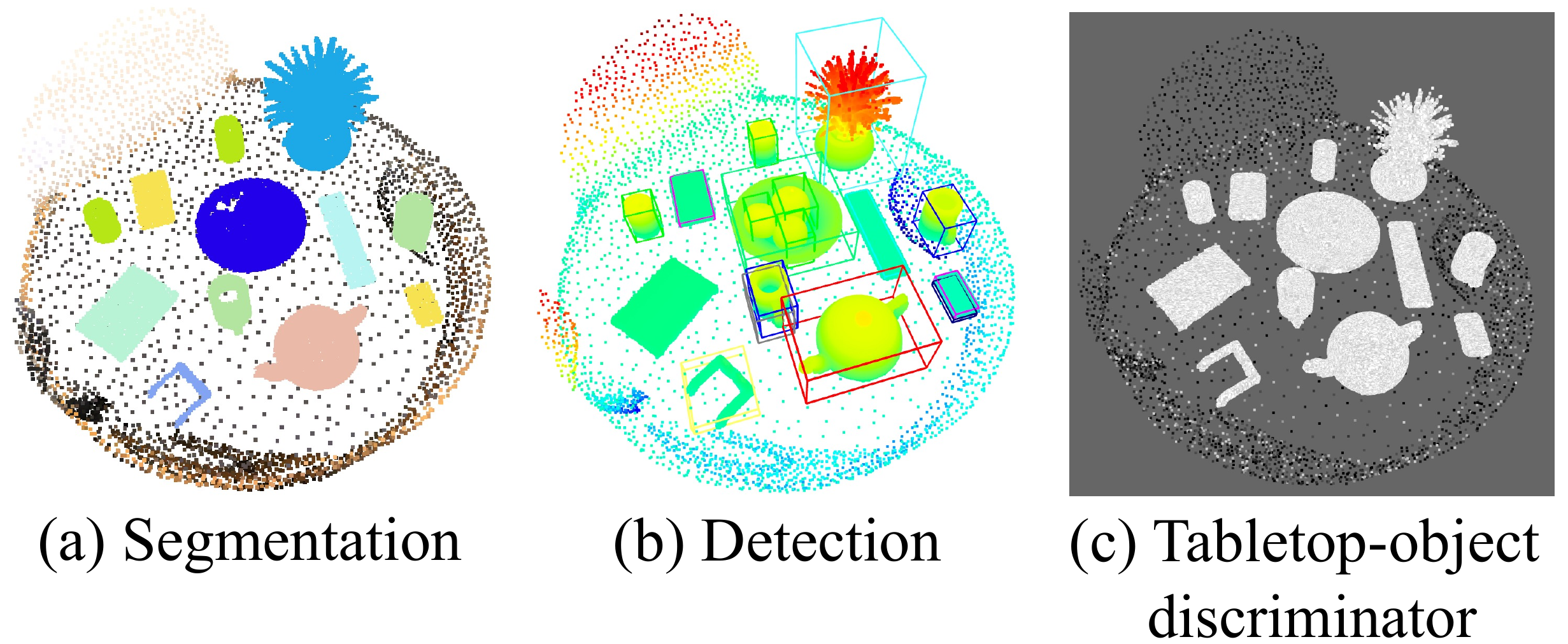}
\caption{Benchmark result.}
\label{fig:task_result}
\end{minipage}
\begin{minipage}[t]{0.35\linewidth}
\centering
\includegraphics[height=3.0cm]{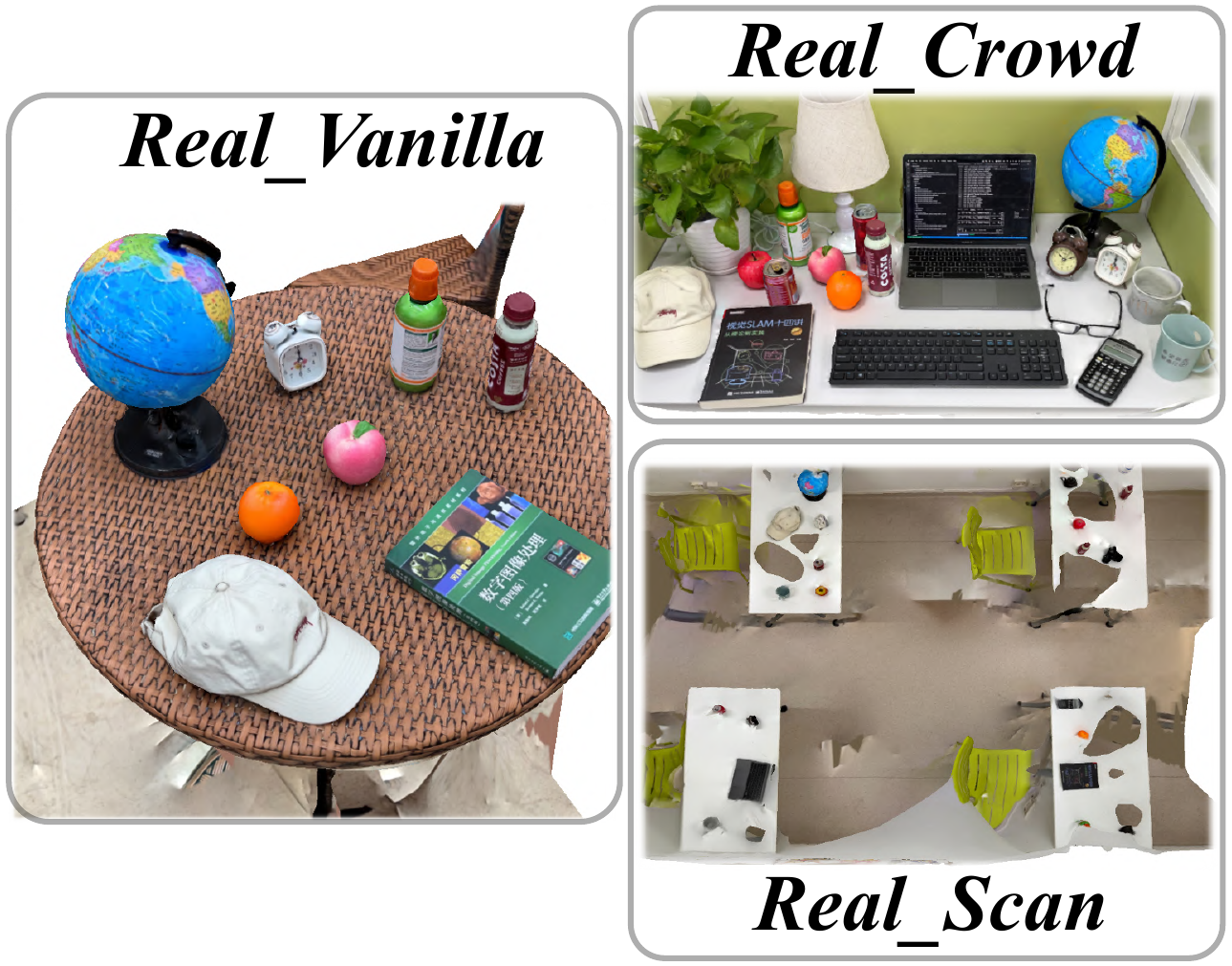}
\caption{TO-Real test set.}
\label{fig:real_test}
\end{minipage}
\end{figure}


A sample of segmentation, detection result and the predicted tabletop-belongings is visualized in Fig. \ref{fig:task_result}. We can see that the networks successfully segment or detect the objects with tabletop-awareness.
Both the segmentation and detection results show that TO\_Crowd is more challenging than TO\_Vanilla, and TO\_ScanNet is most difficult.
More result visualizations, the result of each class, and the ablations of tabletop-aware learning strategies are presented in the supplementary material.


\section{Real-world Test}
The ultimate target of our dataset is for serving the real applications. For a clearer picture of this goal, the below steps are performed.~\\

\noindent{\textbf{Data.}}
Since there is no real dataset that perfectly match with the three variants of TO-Scene, the first thing is to acquire real-world data.
We employ expert users to manually scan and annotate \textbf{TO-Real} including three sets of data (see Fig. \ref{fig:real_test}), which are respectively denoted as Real\_Vanilla, Real\_Crowd, Real\_Scan. Specifically, Real\_Vanilla contains 97 tabletop scenes, while Real\_Crowd includes 100 tabletop scenes with crowded objects. In Real\_Scan, 22 scenes are scanned with both big furniture and small tabletop objects. The categories in TO-Real cover a subset of our TO-Scene objects (see Table \ref{table:real_test}).~\\

\noindent{\textbf{Implementations.}}
We train Point Transformer \cite{pointtrans} for semantic segmentation and VoteNet \cite{votenet} for detection on different variants of TO-Scene, and \textit{directly} test on the corresponding TO-Real counterparts, without any fine-tuning.~\\

\noindent{\textbf{Results and Analysis.}}
Table \ref{table:real_test} enumerates the segmentation mIoU and detection mAP of each class, where we find \textbf{\textit{two important phenomenons}}:

(1) Under both Vanilla and Crowd settings that \textit{specifically} for parsing tabletop objects, the variance between the results tested on TO-Scene and TO-Real is stably acceptable. For detection task, the test mAP (highlighted in red bold) of some categories on TO-Real is even better than it on TO-Scene. This definitely proves the practical value of the \textit{tabletop}-scenes data in TO-Scene.

\textbf{(2)} The more interesting case lies on whole-room Scan (emphasized in gray shadow). As for the big furniture categories (highlighted in blue), the result variance of these categories are obviously undesirable. Note that these categories in TO-Scene all come from ScanNet \cite{scannet}. Additionally, the results of tabletop object classes \textit{particularly degrade} under whole room Scan setup, which is probably influenced by ScanNet big furniture that are simultaneously learned with tabletop objects during training on TO\_ScanNet. 

Here we discuss some possible reasons. Scanning big furniture around large room spaces is physically hard to control, causing various unstable noise (point density, \textit{incompleteness}), and instance \textit{arrangements/layouts} greatly change geographically. These unavoidable factors possibly cause the poor model generalization on big furniture across various data collections. As for tabletop objects, we guess the domain gap is small because scanning tabletops is physically controllable, bringing less unstable noise.

More details of TO-Real and the result visualizations are illustrated in the supplementary material.

\begin{table}[t]
\centering
    \caption{Per-category test results on TO-Scene\textbf{/}TO-Real.}
        \label{table:real_test}
        \setlength{\tabcolsep}{2.0pt}
        \resizebox{0.85\textwidth}{!}{
        \begin{tabular}{l|cc>{\columncolor{gray!20}}c||cc>{\columncolor{gray!20}}c}
        \Xhline{1.0pt}
          Class in & \multicolumn{3}{c||}{Segmentation mIoU (\%)} & \multicolumn{3}{c}{Detection mAP@0.25(\%)}\\
 Real Data & Vanilla & Crowd & Scan & Vanilla & Crowd & Scan\\
 \Xhline{0.6pt}
        \multicolumn{7}{l}{\textcolor{blue}{\textbf{Big furniture:}}}\\ 
        \Xhline{0.6pt}
        \textcolor{blue}{wall} &  - & - & 76.8/10.0 & - & - & - \\
        \textcolor{blue}{floor} &  - & - & 94.9/46.5 & - & - & - \\
        \textcolor{blue}{cabinet} &  - & - & 59.5/19.3 & - & - & 78.5/3.8 \\
        \textcolor{blue}{chair} &  - & - & 80.4/33.6 & - & - & 92.3/$55.7$ \\
        \textcolor{blue}{sofa} &  - & - & 75.3/39.2 & - & - & 98.7/62.2 \\
        \textcolor{blue}{table} & - & - & 71.9/23.5 & - & - & 83.4/$16.3$ \\
        \textcolor{blue}{door} &  - & - & 55.6/19.2 & - & - & 69.5/10.1  \\
        \textcolor{blue}{window} &  - & - & 59.5/5.9 & - & - & 57.1/1.5 \\
        \textcolor{blue}{bookshelf} &  - & - & 64.0/10.2 & - & - & 69.6/10.5 \\
        \textcolor{blue}{picture} &  - & - & 20.8/21.8 & - & - &  16.7/18.5 \\
        \textcolor{blue}{counter} &  - & - & 58.5/6.2 & - & - & 86.0/8.7 \\
        \textcolor{blue}{desk} &  - & - & 62.8/4.1 & - & - & 95.2/3.3\\
        \textcolor{blue}{curtain} &  - & - & 58.3/4.6 & - & - & 76.6/41.7 \\
        \textcolor{blue}{refrigerator} & - & - & 61.8/1.2 & - & - & 93.9/21.3 \\
        \textcolor{blue}{sink} &  - & - & 58.1/34.0 & - & - & 67.9/$70.0$ \\
        \Xhline{0.6pt}
        \multicolumn{7}{l}{\textbf{Tabletop object:}}\\ 
        \Xhline{0.6pt}
        bottle  &  88.3/62.5 & 87.9/70.3 & 85.3/23.6 & \textcolor{red}{64.3/\textbf{67.3}} & 72.4/49.3 & 91.4/33.9 \\
        bowl & 89.5/61.1 & 87.5/51.2 & 85.7/1.8 & 75.5/45.7 & 77.0/69.2 & 90.4/49.2 \\
        camera & 89.0/74.4 & 85.3/77.9 & 81.2/1.2 & \textcolor{red}{81.0/\textbf{91.7}} & 74.7/50.5 & 95.2/3.1 \\
        cap & 92.1/64.2 & 87.8/63.1 & 85.4/28.0 & 87.1/62.8 & 88.2/71.7 & 97.8/$40.0$ \\
        keyboard &  86.0/75.7 & 89.3/77.1 & - & \textcolor{red}{51.2/\textbf{67.4}} & 37.7/36.1 & - \\
        display &  93.6/81.4 & 91.4/80.6 & - & 93.9/92.14 & 84.8/80.2 & - \\ 
        lamp & 86.0/72.6 & 90.0/78.4 & 72.2/10.9 & 84.7/84.2 & 85.6/67.5 & 98.4/61.6 \\
        laptop & 94.3/81.6 & 96.9/66.2 & 95.5/52.7 & 90.3/62.8 & 96.8/69.2 & 97.7/80.1 \\
        mug & 94.2/48.8 & 96.4/72.0 & 92.0/31.2 & 81.8/77.3 & 90.1/$67.0$ & 94.6/32.1 \\
        alarm & 66.7/52.2 & 66.0/54.0 & 51.5/2.7 & 59.2/32.5 & 48.9/38.9 & 94.7/9.5 \\
        book & 77.5/58.6 & 67.3/41.5 & 68.6/28.2 & 60.0/57.2 & \textcolor{red}{62.7/\textbf{69.8}} & 92.8/11.8 \\
        fruit & 94.1/47.4 & 88.7/42.1 & 81.6/16.6 & 76.1/55.9 & 77.9/44.9 & 88.8/31.1 \\
        globe & 96.0/85.1 & 98.0/75.4 & 96.4/17.6 & \textcolor{red}{87.9/\textbf{88.6}} & 95.9/91.1 & 99.9/31.1 \\
        plant & 93.9/65.0 & 96.9/65.6 & 92.0/21.0 & \textcolor{red}{87.7/\textbf{89.2}} & 89.7/80.7 & 97.2/13.3 \\
        \Xhline{1.0pt}
        \end{tabular}
        }
    \label{real_test}
\end{table}

\section{Discussion and Conclusion}
This paper presents TO-Scene, a large-scale 3D indoor dataset focusing on tabletop scenes, built through an efficient data acquisition framework. Moreover, a tabletop-aware learning strategy is proposed to better discriminate the small-sized tabletop instances, which improve the state-of-the-art results on both 3D semantic segmentation and object detection tasks. A real scanned test set, TO-Real, is provided to verify the practical value of TO-Scene. 
One of the variants of our dataset, TO-ScanNet, includes totally 4663 scans with 137k instances, which can possibly serve as a platform for \textit{pre-training} data-hungry algorithms in 3D tasks towards individual shape-level or holistic scene-level.~\\

\noindent{\textbf{Acknowledgements.}}
The work was supported in part by the National Key R\&D Program of China with grant No. 2018YFB1800800, the Basic Research Project No. HZQB-KCZYZ-2021067 of Hetao Shenzhen-HK S\&T Cooperation Zone, by Shenzhen Outstanding Talents Training Fund 202002, by Guangdong Research Projects No. 2017ZT07X152 and No. 2019CX01X104, and by the Guangdong Provincial Key Laboratory of Future Networks of Intelligence (Grant No. 2022B1212010001). It was also supported by NSFC-62172348, NSFC-61902334 and Shenzhen General Project (No. JCYJ20190814112007258). We also thank the High-Performance Computing Portal under the administration of the Information Technology Services Office at CUHK-Shenzhen.

\clearpage
%
%

\bibliographystyle{splncs04}


\clearpage
\begin{appendices}
\section*{\Large \textit{Supplementary Material} for TO-Scene}
\counterwithin{figure}{section}
\counterwithin{table}{section}
\renewcommand{\thesection}{\Alph{section}}%

\centerline{\large{\textbf{Outline}}}
\noindent{This supplementary document is arranged as follows:}\\
(1) Sec.~\ref{sec:toscene_detail} provides more statistics, sample visualizations and discussions of TO-Scene dataset;\\
(2) Sec.~\ref{sec:acquisition_detail} describes our data acquisition framework more comprehensively;\\
(3) Sec.~\ref{sec:task_detail} elaborates the implementations and ablations for benchmarking on our TO-Scene;\\
(4) Sec.~\ref{sec:toreal_detail} illustrates the details of TO-Real test set and visualizes the test results on it.

\section{TO-Scene Dataset}\label{sec:toscene_detail}

\begin{figure}[htbp]
\centering
\includegraphics[width=\textwidth]{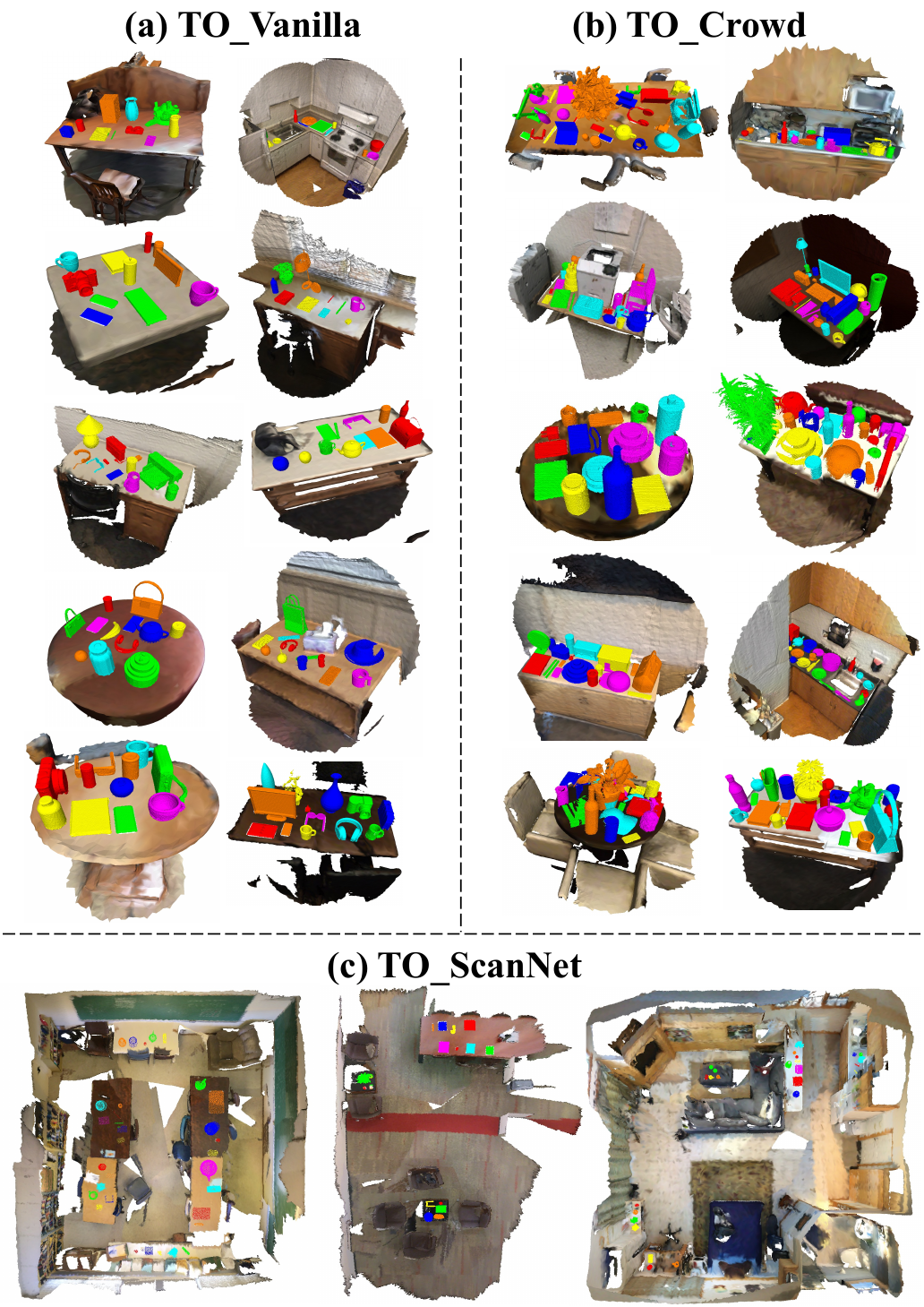}
\caption{A variety of example scenes in three variants of TO-Scene. Each object instance shown with a different randomly assigned color.}
\label{fig:more_samples}
\end{figure}

In this section, we provide detailed statistics and visualize more examples of three variants in TO-Scene, and discuss the current limitations of TO-Scene.

\subsection{Per-category Statistics}
Table \ref{table:per_cat_result} enumerates the train/val split statistics of the instance quantities, plus semantic segmentation mIoU (seg) and object detection mAP@0.25 (det) of each class in three variants of TO-Scene, where our TO-Scene contains a rich amount of instances under rational train/val split, and diverse tabletop object categories in all variants.
For the description of the benchmark results, see Sec.~\ref{sec:task_detail_result}.

\subsection{Example Visualizations}
Fig. \ref{fig:more_samples} presents a larger set of examples in three variants of TO-Scene. For each sample, the surface mesh and separate object instance labels are illustrated. 
It clearly shows the difference across three variants, as well as the outstanding diversity of tabletop objects. In TO\_Crowd, more occlusions and reconstruction errors can be seen (e.g., the right column in the third row of TO\_Crowd).

\subsection{Discussions}
There are two major limitations of the current TO-Scene dataset:

\textbf{(1)} The \textit{texture} of the tabletop object is currently not available due to the lack of the high-quality textured object CADs, but we believe our accessible Web UI opens the chance to replenish our dataset, or build similar datasets for different applications.

\textbf{(2)} \textit{Stacking, collision and suspension} are very common in real-world tabletop scenes. For instance, toothbrushes and toothpastes are in cups, cables and wires are connected to monitors and cellphones, and they are placed randomly on tabletops and colliding with other objects. These scenarios are hard to be designed by our current framework. However, compared with our acquisition method, manually scanning and labelling real tabletop objects with stacking and collision will become \textit{more harsh}. Moreover, our framework firstly makes it possible to build a large-scale dataset of tabletop scenes, through a new insight of “\textit{mixing objects and scenes} by crowd-sourcing”. Inspired by TO-Scene, future works may design to improve our scalable framework and UI, to create more complex scenes realizing stacking, collision or suspension.

\section{Acquisition Framework}\label{sec:acquisition_detail}
This section states more details for speciﬁc steps in our data acquisition framework.

\subsection{Transfer UI}

\begin{figure}[htbp]
\centering
\begin{minipage}[t]{1.0\linewidth}
\centering
\includegraphics[width=\textwidth]{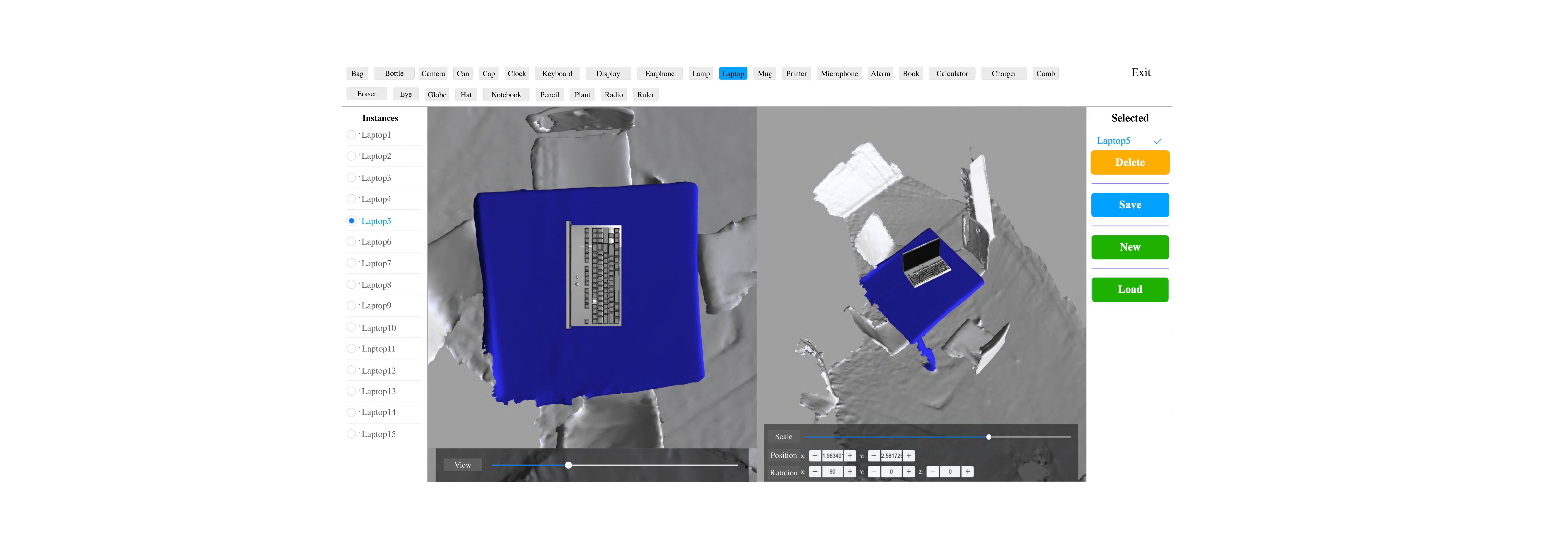}
\caption{Our web-based crowdsourcing interface for transferring CADs onto real tabletops, where the user chooses ``Laptop'' from the category list on the top, then pick ``Laptop5'' from ``Instances'' column at left, and click somewhere on Bird's-Eye-View (BEV) (left) for placing it above 3D tables (right).}
\label{fig:UI_oneobj}
\vspace{1.5cm}
\end{minipage}
\begin{minipage}[t]{1.0\linewidth}
\centering
\includegraphics[width=\textwidth]{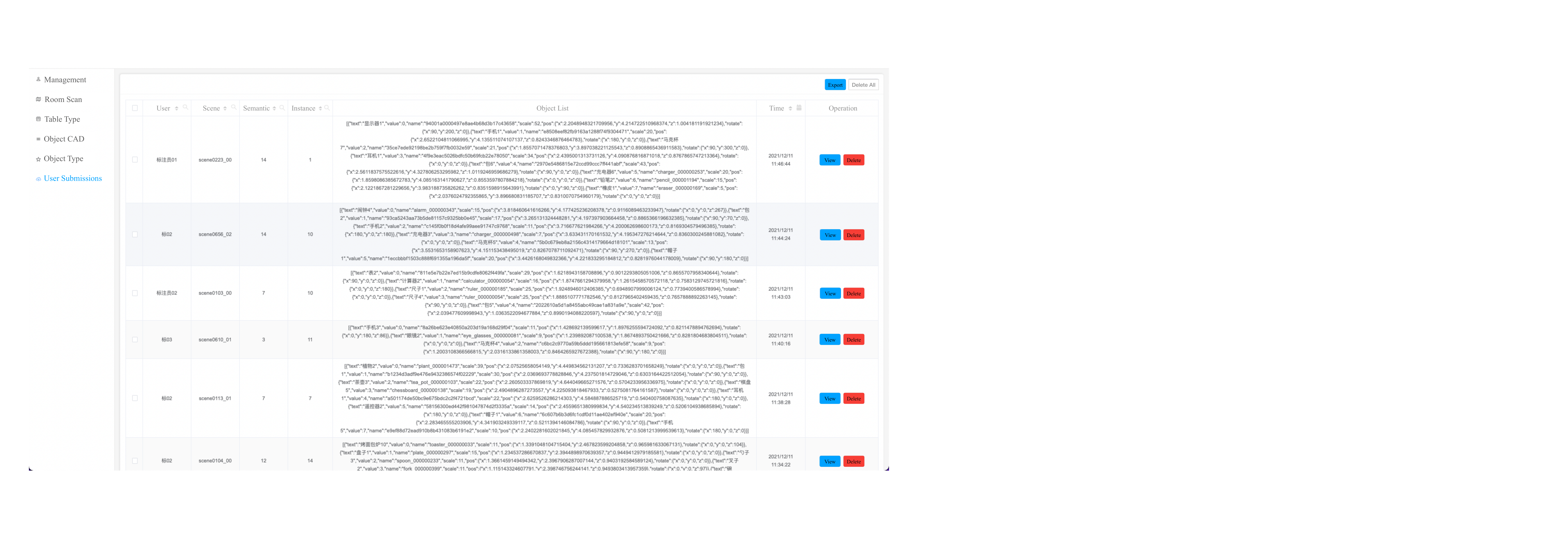}
\caption{The backend management system in our UI, where the user submission interface is shown here.}
\label{fig:UI_backend}
\end{minipage}
\end{figure}


\footnotetext[1]{\label{footnote:threejs} \url{https://threejs.org}}
\footnotetext[2]{\label{footnote:nodejs} \url{https://nodejs.org}}

We use two open-source platforms, Three.js $^{\ref{footnote:threejs}}$ and Node.js $^{\ref{footnote:nodejs}}$, to develop an easy-to-use Web UI to achieve the object transfer process. As shown in Fig. \ref{fig:UI_oneobj}, the operation is basically consisted of a few clicks on UI.~\\

\noindent{\textbf{Source load and selection.}}
At the beginning of each transfer, when an user presses the button ``Load'' at the lower right corner to load new data, a table will be automatically picked from ScanNet \cite{scannet} and loaded. Then a Bird's-Eye-View (BEV) of the loaded table will be shown at left, accompanied with a 3D view of its surroundings at right. At the same time, the semantic classes of small objects are listed at the top of the page, which are also randomly selected from suitable classes in ModelNet \cite{modelnet} and ShapeNet \cite{shapenet} matched with the type of the table (pre-defined in UI according to commonsense knowledge). An user may observe the context around the table and choose one suitable category of the objects, and the ``Instances'' column will be provided at left for picking an object instance of the chosen class.~\\

\noindent{\textbf{Object placement.}}
Notably, when placing an object, the user does \textit{not} need to perceive the \textit{height} to place it in 3D. The only thing to do is just clicking at somewhere on the table in BEV, then the selected object will be automatically placed on the position as the user suggests. We integrate a precise algorithm inside the UI to achieve this, where the height is calculated by the maximum height of the local region on the table within the suggested location to place the object, which ensures the object is transferred as precise as possible, with no gap nor intersection between the table and object.~\\

\noindent{\textbf{Transfer completion.}} 
Simultaneously with each click on BEV, the produced object placement will also be shown in 3D scenes at right, where user is able to drag the 3D scene at right to check the transferred object from comprehensive views, so that its position is able to be adjusted by some simple re-clicks on BEV. Moreover, the toolbar listed at the bottom of the page assist the users to \textit{fine-tune} the scale, position and orientation (operated by modifying the pitch, yaw and roll angles) of the object, arranging the objects in a way that closer to real life scenarios. After finish transferring, the calibration, alignment and other parameters that can re-realize the final transferred result are all saved in a configuration file indicated by the sudo-mappings between different tables and CADs. The average transfer time on a table per crowd worker is 1.8 min for TO\_Vanilla, and 4.5 min for TO\_Crowd.~\\

\noindent{\textbf{Backend management.}} 
To enable high scalability of our data acquisition framework, and continual transparency into the transfer progress, we also endow our UI with a backend management system to track and organize the data (see Fig. \ref{fig:UI_backend}). The left column provides the icon to manage (i.e., revise or delete) the source data of room scans paired with table categories \cite{scannet}, and object CADs coupled with object classes \cite{modelnet,shapenet}, as well as the user submissions of transferred tabletop scenes. This management significantly guarantees the data quality.

\subsection{Real-scene Simulation}
To eliminate the domain gap between synthetic tabletop objects and real-world scans, we simulate the CADs into realistic data.

We run Blender \cite{blender} on several CPUs to render CAD objects into averaged 50-100 RGB-D sequences, with mean time 9.8 sec on each tabletop scene.

\footnotetext[3]{\label{footnote:tsdf} \url{https://github.com/andyzeng/tsdf-fusion-python}}

We select average 26.5 RGB-D frames to reconstruct each scene by implementing TSDF fusion \cite{tsdf} $^{\ref{footnote:tsdf}}$ in a multiprocessing manner, where the mean reconstruction time per sample is 50 sec.

\section{Benchmark Tasks}\label{sec:task_detail}
This section presents more experimental details and ablations for benchmarking the state-of-the-arts on our dataset.

\subsection{Implementation}
As described in the main paper, we follow the original training strategies and data augmentations of all selected methods. Commonly, we run different deep networks using PyTorch \cite{pytorch} via several Nvidia GeForce RTX 3090 GPUs.

When running experiments on TO\_Vanilla and TO\_Crowd, the semantic labels (52 categories) of tabletop objects in our dataset are \textit{all} recognized during training for thorough exploitation.
For TO\_ScanNet, besides keep using all 52-classes labels on tabletop objects, we use NYU2 labels \cite{nyuv2} on the big furniture extracted from ScanNet \cite{scannet}, following the original settings in ScanNet.

According to \cite{scannet} and the state-of-the-arts \cite{paconv,pointtrans,h3dnet,groupfree}, we do not use RGB input. Additionally, for parsing tabletop scenes of TO\_Vanilla and TO\_Crowd, taking table classes in one-hot format as an additional input prior may help learning, but here we ignore it for simplification.

\subsection{Ablations of Tabletop-aware Learning.}
The tabletop-aware learning strategy is proposed to tackle the difficulty of perceiving the small tabletop objects under indoor environments, which is especially challenging on TO\_ScanNet with lots of big furniture surroundings.
Thus, we explore more about our tabletop-aware learning strategy on TO\_ScanNet for 3D semantic segmentation task.~\\

\begin{figure}[htbp]
\centering
\begin{minipage}[t]{0.49\linewidth}
\centering
\captionof{table}{Ablations on $\lambda$.}
\label{table:ablation_lambda}
\resizebox{0.8\textwidth}{!}{
\begin{tabular}{c|c}
\Xhline{1.0pt}
\noalign{\smallskip}
$\lambda$ & Segmentation mIoU (\%)\\
\noalign{\smallskip}
\Xhline{0.6pt}
\noalign{\smallskip}
0.0 (Baseline) & 67.17\\
\noalign{\smallskip}
1.0 & 66.12\\
\noalign{\smallskip}
0.1 & 67.96\\
\noalign{\smallskip}
0.01 & \textbf{69.09}\\
\noalign{\smallskip}
0.001 & 67.58\\
\Xhline{1.0pt}
\end{tabular}}
\end{minipage}
\begin{minipage}[t]{0.49\linewidth}
\centering
\captionof{table}{Ablations on sampling.}
\label{table:ablation_ds}
\resizebox{0.8\textwidth}{!}{
\begin{tabular}{l|c}
\Xhline{1.0pt}
\noalign{\smallskip}
Sampling & Segmentation mIoU (\%)\\
\noalign{\smallskip}
\Xhline{0.6pt}
\noalign{\smallskip}
FPS (Baseline) & 67.17\\
\noalign{\smallskip}
grid & 67.18\\
\noalign{\smallskip}
random & 65.52\\
\noalign{\smallskip}
dynamic (Ours) & \textbf{68.34}\\
\noalign{\smallskip}
\Xhline{1.0pt}
\end{tabular}}
\end{minipage}

\vspace{0.5cm}
\begin{minipage}[t]{0.48\linewidth}
\centering
\includegraphics[height=2.3cm]{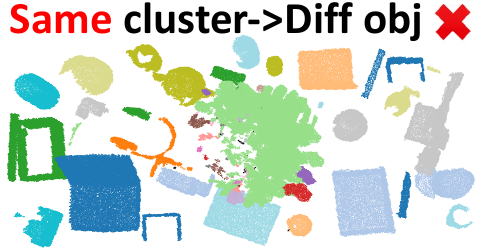}
\caption{Cluster result.}
\label{fig:cluster}
\end{minipage}
\begin{minipage}[t]{0.48\linewidth}
\centering
\includegraphics[height=2.3cm]{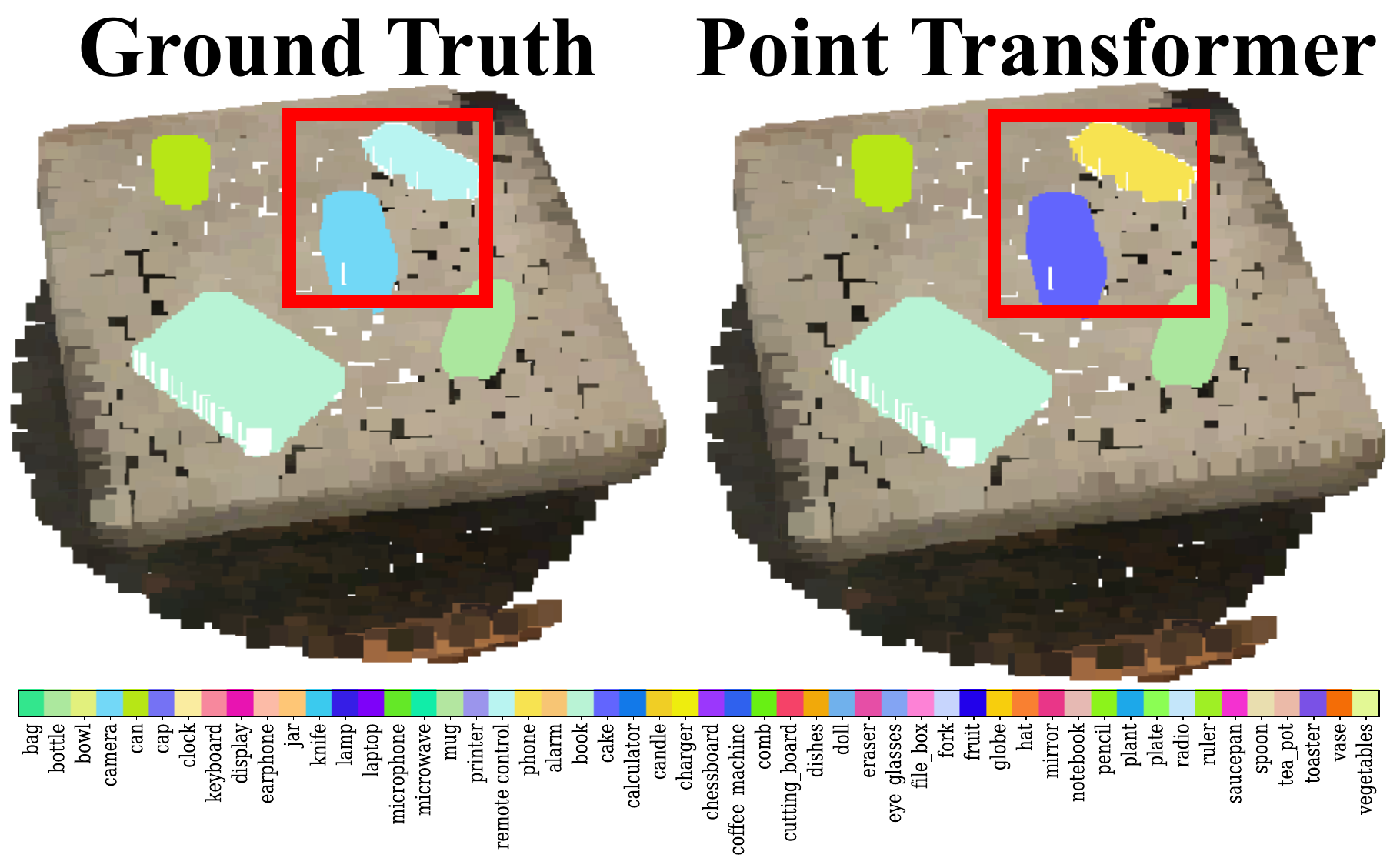}
\caption{A failure case.}
\label{fig:failure}
\end{minipage}

\vspace{0.5cm}
\begin{minipage}[t]{1.0\linewidth}
\centering
\includegraphics[width=\textwidth]{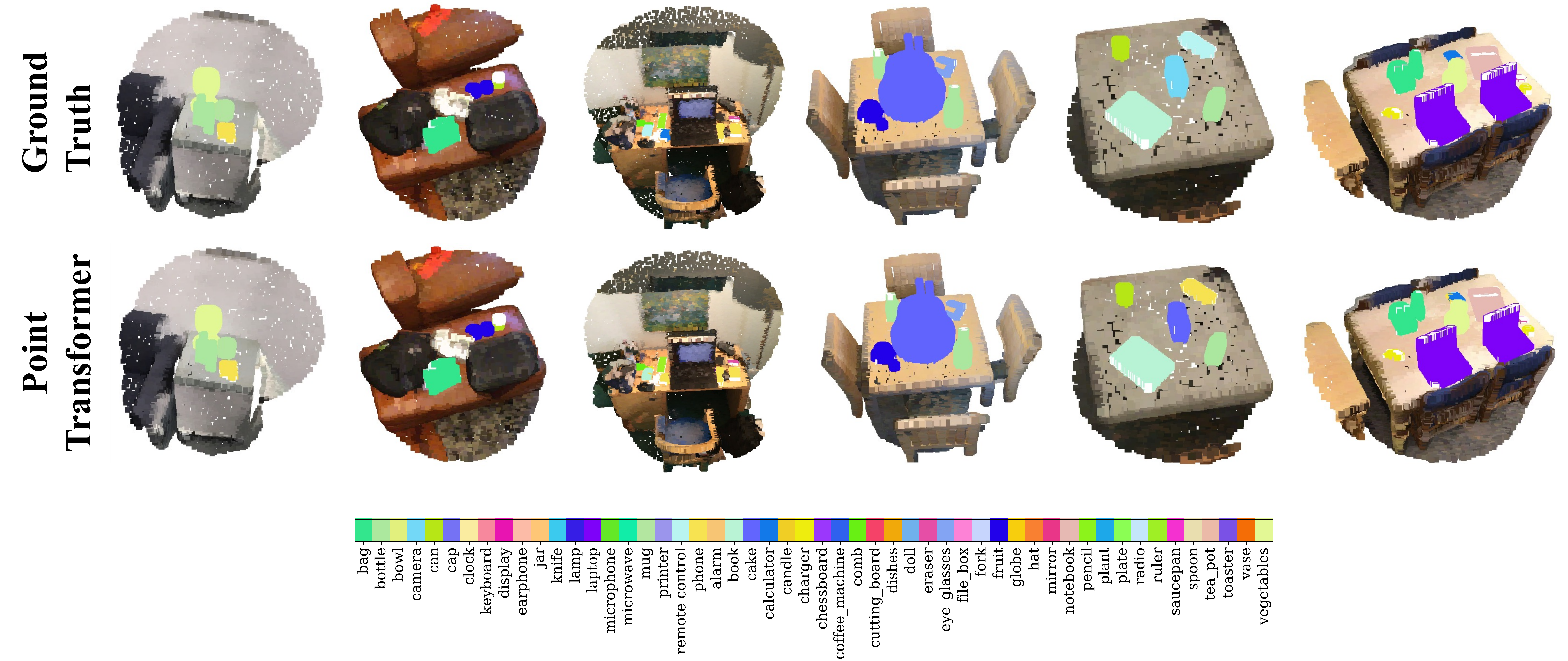}
\caption{Visualization of 3D semantic segmentation results on our dataset. The first row is the ground truth, and the second row is the scenes segmented by Point Transformer \cite{pointtrans}. Each column indicates a scene. The colors in the bottom colorbar represent categories consistently across all scenes.}
\label{fig:vis_seg}
\end{minipage}

\vspace{0.5cm}
\begin{minipage}[t]{1.0\linewidth}
\centering
\includegraphics[width=\textwidth]{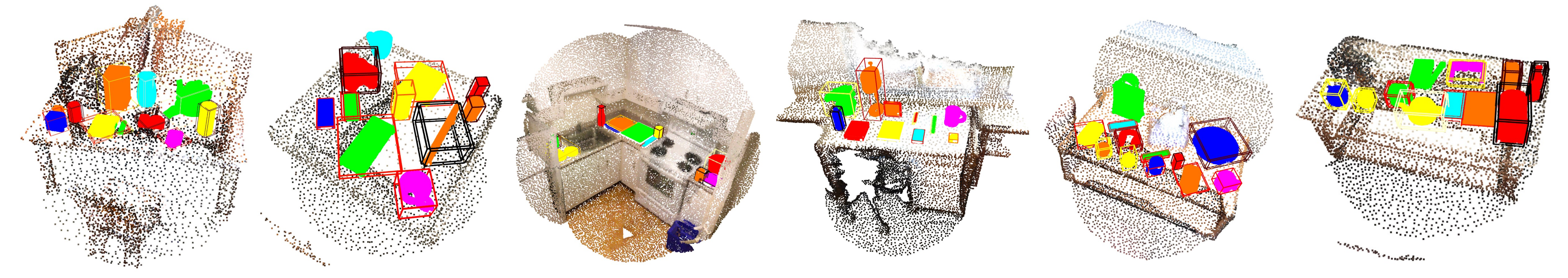}
\caption{Visualization of 3D object detection results by VoteNet \cite{votenet} on our dataset. Each column denotes a scene. Each object instance shown with a different randomly assigned color.}
\label{fig:vis_det}
\end{minipage}
\end{figure}


\noindent{\textbf{$\lambda$ on $L_{dis}$.}} 
As stated in Sec. 4 of the main paper, by introducing tabletop-object discriminator, the loss can be written as: $L_{total} = L_{seg\,or\,det}+ \lambda L_{dis}$, where $\lambda$ is the weight on the discriminator loss $L_{dis}$. We apply feature vector of tabletop-object discriminator on Point Transformer \cite{pointtrans}, and conduct experiments under different $\lambda$. As listed in Table \ref{table:ablation_lambda}, the model performs best when choosing 0.01 as $\lambda$, which provides a decent learning balance. The result also degrades under 1.0 $\lambda$, which may caused by the over optimization on the tabletop-object discriminator.~\\

\noindent{\textbf{Sampling methods.}} 
A derived dynamic sampling is proposed in the main paper.
To make a more complete comparison, we also employ random sampling and grid sampling (i.e., voxelization) on Point Transformer \cite{pointtrans}. Table \ref{table:ablation_ds} summarizes the results, where our dynamic sampling improves baseline with 1.17$\%$\(\uparrow\). Grid sampling has no obvious effect, while random sampling hurts the performance.~\\

\noindent{\textbf{Tasks are still challenging.}}
Although under the help of our tabletop-aware learning strategies, the tasks are still difficult because:

(\textbf{1})~\textit{Discriminator itself is challenging}: Discriminator is not 100\% accurate especially on the joint areas of objects and tables (Paper~Fig.~7 (b)), and will also misguide the learning of downstream networks, leaving the tasks difficult.

(\textbf{2})~\textit{Clustering is not easy even based on perfect discriminator}: We extract tabletop objects from scenes (i.e., equivalent to perfect discrimination) then use DBSCAN, a popular point cloud cluster method. However, it performs not well due to the squeeze of objects (Fig.~\ref{fig:cluster}). Such failure is naturally expected when introducing big furniture from TO\_ScanNet.

(\textbf{3})~\textit{Segmentation/detection is not easy even based on perfect clustering}: Generally speaking, an ideal cluster algorithm can cluster the points of a same object, but it can NOT infer instance categories. For example in Fig.~\ref{fig:failure}, even applying a powerful deep network on simple TO\_Vanilla for semantic segmentation, two objects are yet \textbf{misclassified} while the instances are successfully clustered.


\subsection{Results.}\label{sec:task_detail_result}
Table \ref{table:per_cat_result} exhaustively itemizes the train/val split statistics of the instance quantities, plus semantic segmentation mIoU (seg) and object detection mAP@0.25 (det) of each category in three variants of TO-Scene.

Moreover, Fig. \ref{fig:vis_seg} and Fig. \ref{fig:vis_det} intuitively visualize the results on our dataset for semantic segmentation and object detection tasks using Point Transformer \cite{pointtrans} and VoteNet \cite{votenet}, respectively. As you can see, our dataset successfully drives the state-of-the-arts to perform stably well on both tasks.

\begin{table}[htbp]
\centering
    \caption{Per-category train/val instance quantities and benchmark results on the three variants of TO-Scene. ``seg'' denotes segmentation mIoU (\%), ``det'' indicates detection mAP@0.25 (\%).}
        \label{table:per_cat_result}
        \setlength{\tabcolsep}{2.0pt}
        \resizebox{!}{0.72\textwidth}{%
        \begin{tabular}{l|c|c|c||c|c|c||c|c|c}
        \Xhline{1.0pt}
          Category & \multicolumn{3}{c||}{TO\_Vanilla} &  \multicolumn{3}{c||}{TO\_Crowd} &  \multicolumn{3}{c}{TO\_ScanNet}\\
 & train/val & seg & det & train/val & seg & det & train/val & seg & det\\
 \Xhline{0.6pt}
        \multicolumn{10}{l}{\textbf{Tabletop object:}}\\ 
        \Xhline{0.6pt}
        bag  & 1.6k/368 & 94.8 & 87.8 & 1.4k/341 & 95.5 & 85.8 & 1.8k/346 & 91.2 & 87.3\\
        bottle  & 3.2k/670 & 90.1 & 64.2 & 2.2k/378 & 87.9 & 72.4  & 3.2k/584 & 85.3 & 77.5\\
        bowl & 694/93 & 94.6 & 75.4 & 550/84 & 87.5 & $77.0$  & 681/84 & 85.7 & 69.8\\
        camera & 881/183 & 87.1 & $81.0$ & 1k/234 & 85.3 & 74.7  & 1k/188 & 81.1 & 81.6\\
        can & 5.3k/1k  & 93.6 & 70.4 & 4.4k/869 & 96.2 & $76.0$  & 6.2k/1k & 90.2 & 82.2\\
        cap & 995/225 & 93.5 & 87.1 & 772/207 & 87.8 & 88.2  & 979/235 & 85.4 & 90.5\\
        clock & 242/40 & 31.1 & 28.6 & 152/39 & 28.9 & 18.8  & 255/45 & 14.6 & 20.5\\
        keyboard & 169/49 & 86.4 & 51.2 & 260/56 & 89.3 & 37.7  & 163/45 & $81.0$ & $39.0$\\
        display & 337/104 & 94.4 & 93.9 & 336/77 & 91.4 & 84.8  & 331/105 & 89.7 & 92.5\\
        earphone & 576/164 & 93.2 & 79.7 & 859/181 & 96.7 & 85.8  & 670/181 & 91.8 & 73.1\\
        jar & 335/47 & 63.9 & 39.5 & 262/43 & 61.7 & 53.5  & 334/36 & 64.6 & $60.0$\\
        knife & 88/14 & 39.1 & $19.0$ & 103/22 & 53.3 & 17.5  & 85/13 & 36.9 & 28.9\\
        lamp & 679/149 & 88.9 & 84.6 & 744/150 & $90.0$ & 85.6  & 670/129 & 72.1 & 83.3\\
        laptop & 973/185 & 94.6 & 90.3 & 969/223 & 96.9 & 96.8  & 946/179 & 95.5 & 91.8\\
        microphone & 26/5 & 5.6 & 40.9 & 21/6 & $0.0$ & 1.3 & 34/6 & 16.3 & $0.9$\\
        microwave & 133/26 & 94.6 & 96.7 & 111/21 & 94.3 & $83.0$ & 119/26 & 87.1 & 93.4\\
        mug & 5.4k/1k & $95.0$ & 81.7 & 4.2k/788 & 96.4 & 90.1 & 5.5k/950 & $92.0$ & 86.6\\
        printer & 198/56 & 89.7 & 94.3 & 163/47 & 87.4 & 88.5 & 211/56 & 84.7 & 76.8\\
        remote control & 467/85 & $49.0$ & 19.4 & 385/85 & 51.7 & 29.4 & 558/86 & 33.5 & 24.4\\
        phone & 4k/815 & 76.4 & 24.6 & 3k/655 & 68.8 & 32.7 & 4.3k/748 & 61.3 & 27.6\\
        alarm & 783/155 & 74.3 & 59.2 & 714/107 & $66.0$ & 48.9 & 743/173 & 51.4 & 71.4\\
        book & 1.5k/354 & 81.6 & $60.0$ & 1.5k/334 & 67.3 & 62.7 & 1.6k/349 & 68.5 & 59.5\\
        cake & 254/40 & 83.4 & 83.6 & 282/66 & 76.5 & 68.2 & 309/38 & 85.7 & 57.1\\
        calculator & 1k/239 & 76.8 & 47.7 & 1.1k/197 & 65.7 & 38.4 & 1.2k/223 & 65.2 & 33.6\\
        candle & 268/66 & 88.6 & 62.1 & 190/35 & 40.7 & 53.4 & 306/62 & 78.9 & 75.8\\
        charger & 1.7k/394 & 88.2 & 34.3 & 1.7k/347 & 85.2 & 36.9 & 2.2k/438 & $72.0$ & 39.9\\
        chessboard & 121/20 & 92.1 & 82.0 & 112/13 & 87.2 & 69.3 & 115/27 & 87.9 & 89.4\\
        coffee machine & 118/20 & 92.2 & 91.5 & 100/27 & 86.8 & 91.2 & 108/16 & 97.4 & 86.5\\
        comb & 1k/239 & 91.4 & 40.2 & 322/51 & $70.0$ & 32.7 & 622/142 & 74.3 & 27.7\\
        cutting board & 126/21 & 89.9 & 70.8 & 102/20 & 96.3 & 41.3 & 108/16 & 75.6 & 50.2\\
        dishes & 231/26 & 92.7 & 53.1 & 241/40 & 96.1 & 70.5 & 213/23 & 88.9 & 88.5\\
        doll & 126/24 & $65.0$ & 50.9 & 49/17 & 60.3 & 61.1 & 121/31 & 13.5 & 27.1\\
        eraser & 1k/258 & 65.8 & 9.7 & 1.5k/298 & 71.4 & 12.1 & 1.3k/291 & $34.0$ & 9.1\\
        eye glasses & 2.3k/502 & 93.4 & $68.0$ & 1.7k/342 & 94.8 & 78.6 & 2.5k/483 & 90.4 & 77.0\\
        file box & 168/55 & $93.0$ & 84.2 & 117/42 & 95.8 & $87.0$ & 198/56 & 92.1 & $87.0$\\
        fork & 357/42 & 34.5 & 31.1 & 361/43 & 38.2 & 16.8 & 375/35 & 37.5 & $6.7$\\
        fruit & 2.7k/466 & 94.2 & $76.0$ & 2.5k/463 & 88.7 & 77.9 &3.2k/449  & 81.6 & 75.3\\
        globe & 371/88 & 98.1 & 87.9 & 264/52 & 98.0 & 95.9 & 375/74 & 96.3 & 94.1\\
        hat & 196/43 & $54.0$  & 76.8 & 160/32 & 39.4 & 52.9 & 202/41 & 36.9 & 67.6\\
        mirror & 90/19 & 59.5 & 43.5 & 43/11 & 73.7 & $20.0$ & 44/12 & 21.8 & 33.3\\
        notebook & 1.5k/331 & 69.7 & 28.4 & 1.4k/321 & 62.4 & 32.7 & 1.6k/313 & 54.5 & 30.5\\
        pencil & 1.1k/278 & 71.9 & 14.5 & 1.5k/344 & 75.8 & 15.1 & 1.3k/289 & 58.5 & $6.9$\\
        plant & 1.5k/306 & $94.0$ & 87.7 & 1.3k/251 & 96.9 & 89.7 & 1.4k/288 & 91.9 & 89.9\\
        plate & 390/52 & 93.8 & 34.6 & 314/57 & 98.2 & 47.7 & 375/38 & 90.4 & 67.6\\
        radio & 288/51 & $52.0$ & 50.8 & 283/47 & 50.5 & 60.1 & 280/37 & 37.4 & 27.4\\
        ruler & 858/190 & 73.4 & 13.5 & 832/16 & 67.4 & $10.0$ & 899/189 & 55.8 & $8.9$\\
        saucepan & 153/23 & 94.2 & 73.5 & 121/25 & 87.1 & 57.5 & 127/16 & 69.9 & 75.9\\
        spoon & 429/58 & 60.4 & 32.6 & 365/56 & 46.9 & 19.9 & 456/43 & 49.5 & 16.9\\
        tea pot & 1.1k/201 & 97.5 & 89.7 & 1k/198 & 95.7 & 93.4 & 1.1k/203 & 94.5 & 83.4\\
        toaster & 93/12 & 83.2 & 66.1 & 62/6 & 45.6 & 41.2 & 91/8 & 86.9 & 46.3\\
        vase & 1.2k/243 & 89.5 & 71.5 & 638/141 & 83.4 & 79.1 & 1k/188 & 71.7 & 79.3\\
        vegetables & 73/11 & 3.89 & 34.9 & 85/12 & $9.0$ &  10.5 & 79/13 & 20.1 & 7.4\\
        \Xhline{1.0pt}
        \multicolumn{10}{r}{Continued at next page}\\
        \end{tabular}%
        }
\end{table}
\addtocounter{table}{-1}
\begin{table}[t]
\centering
    \caption{Per-category train/val instance quantities and benchmark results on the three variants of TO-Scene. ``seg'' denotes segmentation mIoU (\%), ``det'' indicates detection mAP@0.25 (\%). (Continued)}
        \setlength{\tabcolsep}{2.0pt}
        \resizebox{!}{0.3\textwidth}{%
        \begin{tabular}{l|c|c|c||c|c|c||c|c|c}
        \Xhline{1.0pt}
          Category & \multicolumn{3}{c||}{TO\_Vanilla} &  \multicolumn{3}{c||}{TO\_Crowd} &  \multicolumn{3}{c}{TO\_ScanNet}\\
 & train/val & seg & det & train/val & seg & det & train/val & seg & det\\
        \Xhline{0.6pt}
        \multicolumn{10}{l}{{\textbf{Big furniture:}}}\\ 
        \Xhline{0.6pt}
        wall & - & - & - & - & - & - & 19.6k/4.1k & 76.7 & - \\
        floor &  - & - & - & - & - & - & 9.8k/2.1k & 94.8 & - \\
        cabinet &  - & - & - & - & - & - & 6.8k/1.3k & 59.4 & 49.3 \\
        bed &  - & - &  & - & - & - & 943/218 & 80.3 & 88.8 \\
        chair &  - & - & - & - & - & - & 15.9k/3.8k & 86.7 & 82.2 \\
        sofa &  - & - & - & - & - & - & 2.1k/331 & 75.2 & 89.5 \\
        table & - & - & - & - & - & - & 6.5k/1.3k & 71.9 & 61.8 \\
        door &  - & - & - & - & - & - & 6.8k/1.4k & 55.6 &  45.8 \\
        window &  - & - & - & - & - & - & 3.5k/863 & 59.4 &  37.1\\
        bookshelf &  - & - & - & - & - & - & 961/306 & 63.9 & 31.7\\
        picture &  - & - & - & - & - & - & 2.2k/628 & 20.8 & $5.2$\\
        counter &  - & - & - & - & - & - & 1k/165 & 58.5 & 45.9\\
        desk &  - & - &  & - & - & - & 2.9k/567 & 62.7 & 70.6\\
        curtain &  - & - & - & - & - & - & 878/137 & 58.3 & 48.9\\
        refrigerator & - & - & - & - & - & - & 887/166 & 61.8 & 61.2\\
        shower curtain & - & - & - & - & - & - & 349/95 & 71.3 & 70.4 \\
        toilet & - & - & - & - & - & - & 522/121 & 85.9 & 88.6 \\
        sink &  - & - & - & - & - & - & 1.3k/284 & 58.1 & 37.5 \\
        bathtub &  - & - & - & - & - & - & 344/104 & 82.6 & $90.0$ \\
        other furniture & - & - & - & - & - & - & 7.5k/1.5k & $45.0$ & 41.3\\
        \Xhline{1.0pt}
        \end{tabular}%
        }
\end{table}

\section{TO-Real}\label{sec:toreal_detail}
The main paper introduces TO-Real that is scanned from real world for verifying the practical value of our TO-Scene.

\subsection{Data}
Here we present more details of TO-Real dataset.
To achieve the diversity, we employ 52 people of various ages from different professions to place the objects following their daily habits, where half of them are guided to arrange crowded objects. The whole process are taken place at diverse indoor rooms (e.g., kitchen, conference room, bedroom, bathroom, lobby, living room) in several offices or homes. After finishing the object placement, we employ 10 experts to scan the tabletop scenes using Microsoft Kinect \cite{kinect} and annotate the data with both point-wise segmentation and object bounding boxes manually, which producing Real\_Vanilla and Real\_Crowd. Similarly, the experts also manually scan and label the whole rooms holding the tables, yielding Real\_Scan. When collecting Real\_Scan, expert workers may walk close to tabletops and scan the small tabletop objects from different views, avoiding the severely undesirable data quality of tabletop objects.

\begin{figure}[t]
\centering
\includegraphics[width=\textwidth]{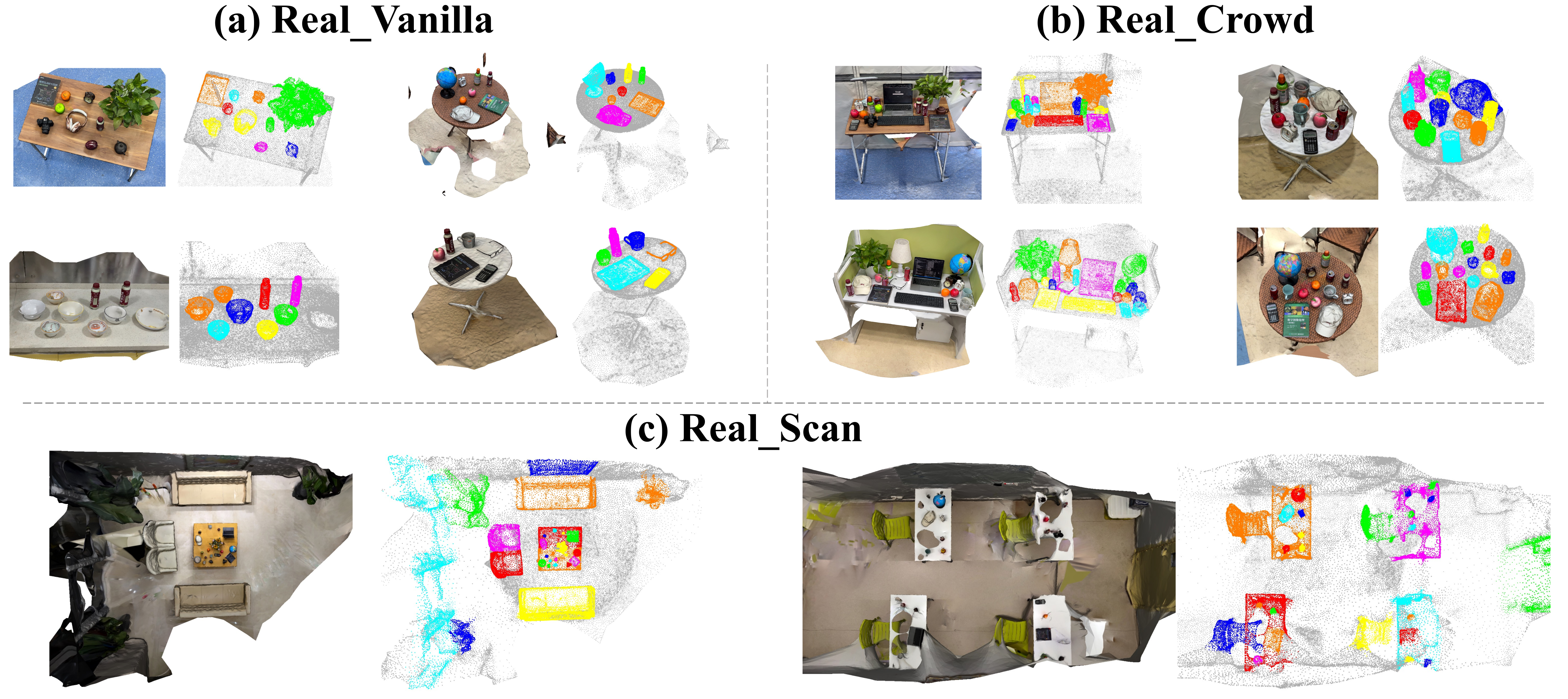}
\caption{A variety of example scenes in three variants of TO-Real. Each sample is shown with the original reconstructed mesh at left, and object instance assigned with random color at right.}
\label{fig:realtest_more}
\end{figure}

Fig. \ref{fig:realtest_more} illustrates more samples in TO-Real. For each reconstruction, the surface mesh with colors is shown, as well as a visualization of separate object instance labels are also available to indicate multiple instances. It conspicuously shows that the characteristics of three variants (Real\_Vanilla, Real\_Crowd, Real\_Scan) exactly match their counterparts in TO-Scene shown in Fig. \ref{fig:more_samples}.

Tab.~\ref{tab:object_distribution} shows the instance distributions of several classes, on which Real\_Scan and TO\_ScanNet are similar, which minimizes the gap between the two datasets.

\begin{table}[htbp]
	\centering
            \begin{tabular}{c|cccccc||cccc}
                \bottomrule[1pt]
                Class & chair & table & window & bookshelf & counter & sink & bottle & camera & cap & fruit \\
    			\hline
    			Real & 32.6 & 9.15 & 6.15 & 1.54 & 1.54 & 1.54 & 10.23 & 3.21 & 3.40 & 11.22\\
    			TO & 27.3 & 10.75 & 6.02 & 1.75 & 1.65 & 2.27 & 9.64 & 3.09 & 3.08 & 9.34\\
                \toprule[0.8pt]
    		\end{tabular}
	\caption{Instance distributions (\%) in Real\_Scan and TO\_ScanNet.}
    \label{tab:object_distribution}
\end{table}

\subsection{Tabletop-aware Learning on TO-Real.}
Tab.~\ref{tab:tabletop_learning} shows the improvements (segmentation mIoU) on several categories after applying the proposed strategies (+FV+DS) in real scans.

\subsection{Visualization of Segmentation.}
\begin{figure}[t]
\centering
\includegraphics[width=\textwidth]{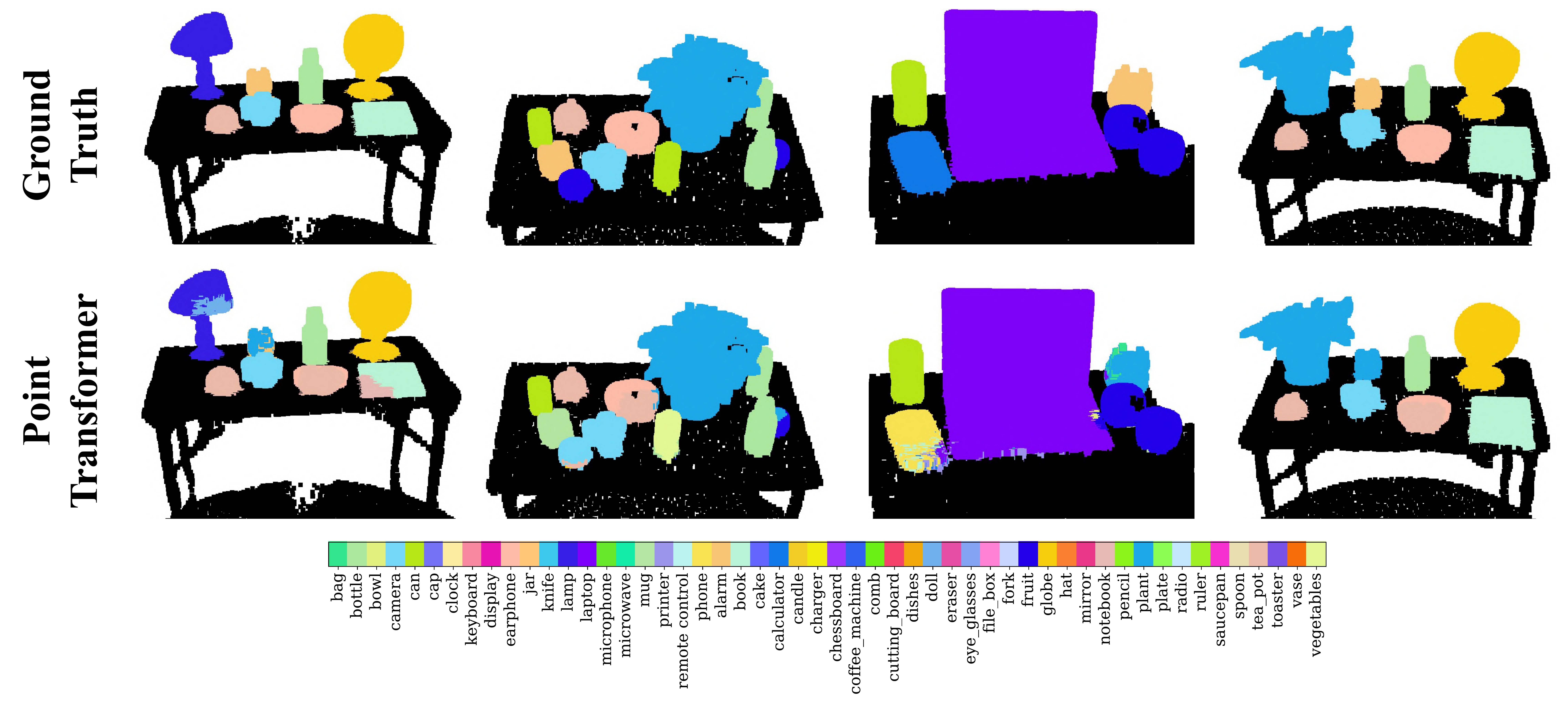}
\caption{Visualization of 3D semantic segmentation test results on TO-Real. The first row is the ground truth, and the second row is the TO-Real scenes segmented by Point Transformer \cite{pointtrans} that trained on our TO-Scene. Each column indicates a scene. The colors in the bottom colorbar indicate classes consistently across all scenes.}
\label{fig:vis_seg_real}
\end{figure}

Fig. \ref{fig:vis_seg_real} visualizes the semantic segmentation result on TO-Real, where Point Transformer \cite{pointtrans} is trained on the training set of TO-Scene, and is \textit{directly} tested on TO-Real. Although purely trained on TO-Scene and tested on realistic data, the model still get acceptable result. This intuitively proves the practical value of our TO-Scene dataset.

\begin{table}[t]
	\centering
            \begin{tabular}{c|ccccccccc}
                \bottomrule[1pt]
                Class & bottle & bowl & cap & keyboard & lamp & mug & alarm & book & fruit \\
    			\hline
    			+FV+DS & 0.43\(\uparrow\) & 0.44\(\uparrow\) & 1.02\(\uparrow\) & 0.54\(\uparrow\) & 0.45\(\uparrow\) & 1.36\(\uparrow\) & 0.74\(\uparrow\) & 0.94\(\uparrow\) & 0.65\(\uparrow\)\\
                \toprule[0.8pt]
    		\end{tabular}
	\caption{Per-category mIoU improvements (\%) on TO-Real.}
    \label{tab:tabletop_learning}
\end{table}

\end{appendices}

\end{document}